\documentclass{article}

\usepackage[preprint]{neurips_2026}

\usepackage[utf8]{inputenc} % allow utf-8 input
\usepackage[T1]{fontenc}    % use 8-bit T1 fonts
\usepackage{hyperref}       % hyperlinks
\usepackage{url}            % simple URL typesetting
\usepackage{booktabs}       % professional-quality tables
\usepackage{amsfonts}       % blackboard math symbols
\usepackage{nicefrac}       % compact symbols for 1/2, etc.
\usepackage{microtype}      % microtypography
\usepackage{xcolor}         % colors

\usepackage{times}
\usepackage{latexsym}
\usepackage{graphicx}
\usepackage{amsmath}
\usepackage{amsthm}
\usepackage{bbm}
\usepackage{caption}
\usepackage{multirow}
\usepackage{subfigure}
\usepackage{subcaption}
\usepackage{pgfplots}
\usepackage{xspace}
\usepackage{enumitem}
\usepackage{wrapfig}
\usepackage{bbding}
\usepackage{colortbl}
\usepackage{amsthm}
\usepackage[ruled,vlined,linesnumbered]{algorithm2e}
\usepackage{sidecap}

\title{Rethinking Experience Utilization in Self-Evolving Language Model Agents}

\author{%
  Weixiang Zhao\textsuperscript{1}, Yingshuo Wang\textsuperscript{1}, Yichen Zhang\textsuperscript{1}, Yanyan Zhao\textsuperscript{1} \\ \textbf{Yu Zhang\textsuperscript{1}}, \textbf{Yang Wu\textsuperscript{2}}, \textbf{Dandan Tu\textsuperscript{2}}, \textbf{Bing Qin\textsuperscript{1}}, \textbf{Ting Liu\textsuperscript{1}}\\
  \textsuperscript{1}Harbin Institute of Technology, \textsuperscript{2}Huawei Technologies Co., Ltd\\
  \texttt{\{wxzhao,yswang,yczhang,yyzhao\}@ir.hit.edu.cn} \\
}

\begin{document}

\maketitle

\begin{abstract}
Self-evolving agents improve by accumulating and reusing experience from past interactions. Existing work has largely focused on how experience is constructed, represented, and updated, while paying less attention to how experience should be used during runtime decision-making. As a result, most agents rely on rigid usage strategies, either injecting experience once at initialization or at every step, without considering whether it is needed for the current decision.
This paper studies experience utilization as a critical design dimension of self-evolving agents. We ask whether agents benefit from interweaving experience use with decision-making, so that experience is invoked only when additional guidance is needed. To examine this question, we introduce {ExpWeaver}, a lightweight instantiation that leaves experience construction unchanged and modifies only runtime utilization by exposing experience as an optional resource during reasoning.
Across four representative frameworks, seven LLM backbones, and three types of environments, ExpWeaver consistently achieves the best performance among different utilization strategies. Reinforcement learning experiments further show that this behavior can be amplified through training. Usage-pattern, causal ablation, and entropy-based analyses reveal that ExpWeaver enables agents to invoke experience selectively, at beneficial decision points, and under higher reasoning uncertainty. Overall, our findings call for a shift from merely studying \emph{what} experience to store toward understanding \emph{how} and \emph{when} experience should enter decision-making.
\end{abstract}

\section{Introduction}

The emergence of self-evolving agents marks a critical milestone in the development of autonomous systems capable of continuous learning and adaptive behavior \citep{zhao2024sapt,dou2025evalearn,fang2025comprehensive}. As we move into the era of experience \citep{silver2025welcome}, where learning is increasingly driven by agents' own interaction histories, this paradigm offers substantial potential for building systems that operate robustly in open-ended environments \citep{hu2025improvisation,lin2026position}.

At a high level, self-evolving agents operate by continuously gathering, storing, and reusing experience from interactions with the environment to inform future decisions \citep{gao2025survey,cai2025building,bell2025future}. This process can be naturally decomposed into two stages: constructing an experience repository and utilizing it during decision-making. Existing research has overwhelmingly focused on the former, with extensive efforts devoted to designing diverse representations of experience, such as distilled insights \citep{zhang2025g,ouyang2025reasoningbank}, reusable skills \citep{wang2024voyager,xia2026skillrl}, or executable workflows \citep{wang2025agent}. In contrast, the question of {how experience should be used at runtime} remains comparatively underexplored.

\begin{figure*}[t]
    \centering
    \includegraphics[width=1\linewidth]{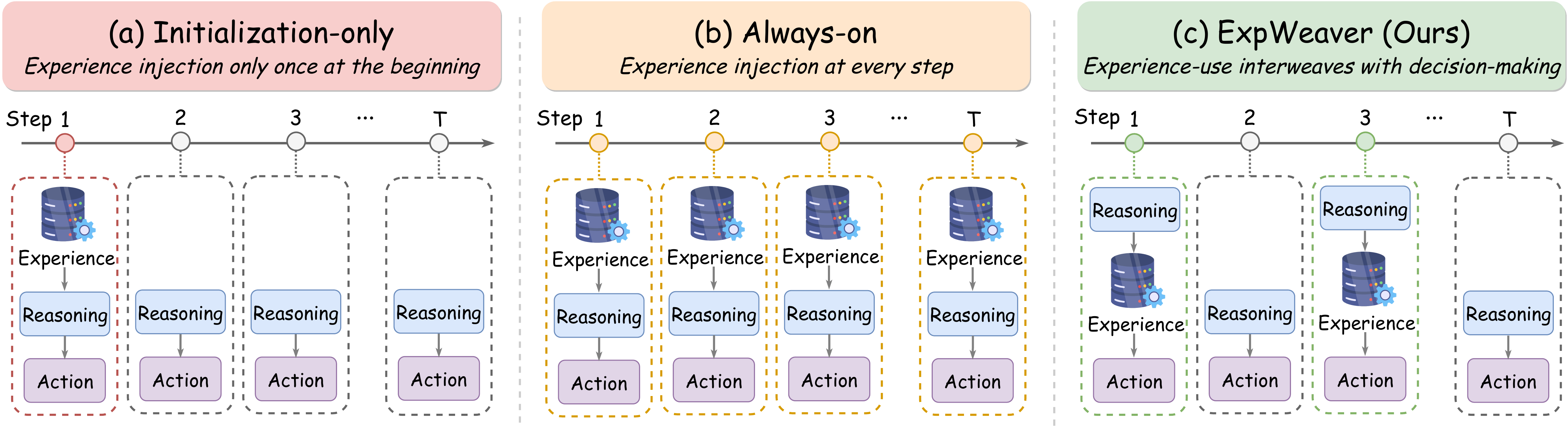}
    \caption{
Illustration of different experience utilization strategies in self-evolving agents. 
(a) \textbf{Initialization-only}: experience is injected once at the beginning and remains static throughout the decision process. 
(b) \textbf{Always-on}: experience is indiscriminately injected at every step, regardless of its necessity. 
(c) \textbf{ExpWeave}: experience utilization is interwoven with the decision-making process, allowing the agent to decide during reasoning {when} to retrieve and use experience.}
    \label{fig:comparison}
\end{figure*}

In practice, as conceptually illustrated in Figure~\ref{fig:comparison}(a) and (b), existing agents rely on rigid experience usage schemes. Some methods follow an \emph{initialization-only} strategy, where experience is injected once at the beginning of task execution \citep{zhao2024expel}; others adopt an \emph{always-on} strategy, where retrieved experience is indiscriminately injected at every decision step \citep{wang2025agent}. Although these strategies are simple and easy to implement, they implicitly assume that experience is either globally useful throughout a task or uniformly useful at every step. However, they do not distinguish whether experience is actually needed for the current decision. This leaves open a fundamental question: {what role should experience play within the agent's runtime decision-making process?}

This gap motivates us to revisit the role of experience in self-evolving agents. In human cognition, experience is not used as static background knowledge, but is tightly coupled with decision-making: individuals selectively apply relevant memories when facing uncertainty, ambiguity, or increased cognitive control demands \citep{shiffrin1977controlled,botvinick2001conflict}. Cognitive neuroscience further suggests that memory retrieval is actively regulated based on task demands, with the prefrontal cortex gating access to stored knowledge during goal-directed behavior \citep{tomita1999top,anderson2004neural,bausch2026distinct}. Inspired by this mechanism, we hypothesize that: \emph{if agents can similarly couple experience utilization with their decision-making process, enabling experience to be invoked only when needed, they may make more effective use of accumulated experience.}

To examine this hypothesis, we propose \textbf{ExpWeaver}, a lightweight paradigm that interweaves experience utilization into the agent's decision-making process. ExpWeaver is not designed to change how experience is constructed, represented, or updated. Instead, it provides a minimal intervention on the usage side: experience is made available as an optional resource during reasoning. Concretely, ExpWeaver extends the widely adopted ReAct-style \citep{yao2023react} decision-making paradigm in self-evolving agents, where decisions are made through iterative \texttt{reasoning} $\rightarrow$ \texttt{action} steps. As illustrated in Figure~\ref{fig:comparison}(c), ExpWeaver introduces a unified loop: \texttt{reasoning} $\rightarrow$ \texttt{(optional experience utilization)} $\rightarrow$ \texttt{action}. In this loop, the agent may retrieve relevant experience only when it determines during reasoning that additional guidance is needed.

This design allows us to systematically study whether regulating experience usage, rather than improving experience representation, is a key missing factor in self-evolving agents. To this end, we integrate ExpWeaver into four representative frameworks with diverse forms of experience, spanning both single-agent \citep{ouyang2025reasoningbank,wang2025agent,xia2026skillrl} and multi-agent settings \citep{zhang2025g}, and compare it against their vanilla experience usage strategies, which follow either initialization-only or always-on schemes. Following the original evaluation settings of these frameworks, we evaluate different experience utilization strategies across seven LLM backbones and three types of environments, including embodied interaction, web navigation, and knowledge-intensive question answering. Across these settings, ExpWeaver consistently outperforms the corresponding vanilla frameworks without modifying existing experience construction pipelines. These results support our central hypothesis that, beyond how experience is constructed and represented, how experience is utilized during decision-making is also a critical factor for self-evolving agents (\S\ref{subsec:prompting_results}).

We further investigate whether interweaving experience utilization with decision-making is merely an inference-time prompting effect or a capability that can be strengthened through learning (\S\ref{subsec:rl_results}). To this end, we incorporate ExpWeaver into agentic reinforcement learning, where experience serves both as inference-time context and as a signal for policy optimization \citep{wu2025evolver,zhai2025agentevolver,xia2026skillrl}. Experiments on Qwen3-4B and Qwen3-14B \citep{yang2025qwen3}, trained with GRPO \citep{shao2024deepseekmath}, show that ExpWeaver consistently achieves the best performance among different experience utilization strategies. This suggests that interweaving experience utilization within decision-making is a learnable capability that can be further amplified through training.

To further understand what interweaving experience utilization with decision-making brings to agents, we conduct a series of analyses that reveal ExpWeaver's behavior and mechanism. First, ExpWeaver induces a non-uniform pattern of experience utilization: agents invoke experience more frequently in demanding tasks and at critical stages of a trajectory, rather than using it uniformly across all steps (\S\ref{subsec:pattern_analysis}). Second, targeted ablations show that this pattern is not incidental but causally beneficial. Removing experience at positions selected by ExpWeaver leads to consistent performance degradation, while random experience utilization fails to yield comparable gains, indicating that the effectiveness of experience depends critically on when it is invoked (\S\ref{subsec:causal_ablation}). Finally, we provide mechanistic evidence that ExpWeaver tends to invoke experience at decision points associated with higher uncertainty, as reflected by increased token-level entropy during the agent's reasoning process (\S\ref{subsec:entropy_analysis}). Together, these findings suggest that coupling experience utilization with decision-making enables agents to adaptively regulate experience use in response to decision difficulty.

Overall, our findings suggest that effective self-evolution depends not only on accumulating better experience, but also on regulating how experience enters the decision-making process. This study highlights experience utilization as a critical design dimension for future self-evolving agents, shifting the focus from merely storing experience to understanding how and when experience should be used.

\section{Related Works}

Existing works on experience-driven self-evolving agents can be broadly discussed from two perspectives: {experience construction} and {experience utilization}.

\emph{Experience construction} studies how agents should represent, store, and update past interactions. A large body of work focuses on constructing external experience repositories from offline datasets or online interactions. Offline approaches pre-construct reusable experience from collected trajectories and keep the memory fixed during inference \citep{li2023mot,yang2023failures,zhong2024memorybank,zhao2024expel,fu2024autoguide,zhou2025memento,yang2025coarse}. Online approaches further allow the experience repository to evolve during deployment, enabling agents to accumulate, retrieve, and refine experience through interaction \citep{chen2024automanual,zhang2025agentic,suzgun2025dynamic,ouyang2025reasoningbank}. Representative systems such as ReasoningBank continuously distill reasoning patterns from recent episodes, while G-Memory extends online memory accumulation to multi-agent settings \citep{ouyang2025reasoningbank,zhang2025g}. These studies have produced diverse experience forms, including distilled insights, skills, workflows, and raw trajectories.

In contrast, \emph{experience utilization} concerns how constructed experience should enter the agent's runtime decision-making process. This aspect remains comparatively underexplored. Recent concurrent studies have begun to explore more adaptive experience usage, such as proactive retrieval in lifelong agents \citep{cai2026ask} and self-triggered experience seeking in web agents \citep{zhang2026expseek}. However, since these works introduce their own experience representations and construction pipelines, the impact of experience utilization is intertwined with changes in how experience is formed, making it difficult to cleanly isolate the effect of \emph{how} experience is used.

Our work instead takes a higher-level view by rethinking the role of experience in self-evolving agents: experience should not merely serve as static auxiliary context, but as an active component of decision-making. To isolate this question, we keep existing experience construction, representation, and update pipelines unchanged, and only modifies runtime utilization. This allows us to identify experience utilization as a critical design dimension, showing that effective self-evolution depends not only on {what} experience is stored, but also on {how} and {when} it enters decision-making.

\section{Interweaving Experience Utilization within Decision-Making}

In this section, we first formulate experience-driven self-evolution as a two-stage process consisting of experience construction and experience utilization (\S\ref{subsec:pre}). We then introduce ExpWeaver, a lightweight intervention that interweaves experience utilization with the agent's reasoning process, enabling us to examine whether experience can be invoked when needed during decision-making (\S\ref{subsec:ExpWeaver}). Finally, we extend this formulation to agentic reinforcement learning to study whether such usage behavior can be further optimized as a learnable capability (\S\ref{subsec:ExpWeaver_rl}).

\subsection{Preliminaries}
\label{subsec:pre}

We study {self-evolving agents}, where behavioral adaptation is achieved by accumulating and reusing past interactions without modifying model parameters \citep{gao2025survey}. This process consists of two stages: constructing an experience repository and utilizing it during decision-making.

\paragraph{Experience Construction.}
After each interaction, the agent produces a trajectory $\tau$ and receives feedback $r$. From each $(\tau,r)$ pair, an {experience unit} $E$ is distilled and stored in an external memory:
\[
M=\{E_1,E_2,\dots,E_n\}.
\]
Different frameworks instantiate $E$ as distilled insights \citep{ouyang2025reasoningbank}, reusable skills \citep{xia2026skillrl}, or executable workflows \citep{wang2025agent}, with corresponding memory update rules. Most works focus on this stage, improving how experience is represented, organized, and maintained.

\paragraph{Experience Utilization.}
Experience utilization determines how the constructed experience memory $M$ enters runtime decision-making, and is largely orthogonal to experience construction. Given a task input $x$, existing methods retrieve relevant experience $M(x)\subseteq M$ and typically use it in fixed and rigid ways. One strategy is \emph{initialization-only} usage:
\[
y=\pi_\theta([x;M(x)]),
\]
where experience is injected once. Another is \emph{always-on} usage in multi-step decision-making:
\[
a_t=\pi_\theta([s_t;M(s_t)]),
\]
where experience is injected at every step $t$.

Both strategies predefine when experience is used, rather than allowing the agent to decide whether it is needed for the current decision. This motivates us to study experience utilization as a runtime decision process: how should experience selectively enter decision-making when useful?

\subsection{Prompting-based ExpWeaver}
\label{subsec:ExpWeaver}

To study this question, we propose \textbf{ExpWeaver}, a lightweight paradigm that interweaves experience utilization with the agent's decision-making process. ExpWeaver leaves experience construction, representation, and updating unchanged, and only modifies runtime utilization by exposing experience as an optional resource during reasoning.

\paragraph{From Fixed Usage to Decision-Time Utilization.}
We build on the widely adopted ReAct-style \citep{yao2023react} paradigm, where agents follow an iterative \texttt{reasoning} $\rightarrow$ \texttt{action} loop. At step $t$, the agent maintains a context $c_t$ and generates an action $a_t \sim \pi_\theta(\cdot \mid c_t)$.

In existing agents, experience utilization is external to this loop: experience is injected either before execution or at every step. ExpWeaver instead places experience utilization inside the \texttt{reasoning} process, making the decision of whether to use experience conditioned on the agent's reasoning state.

\paragraph{Interweaving Experience into Reasoning.}

ExpWeaver implements this idea through a minimal prompting-based intervention. We augment the system prompt with an instruction that reminds the agent that it may consult past experience during reasoning if additional guidance is needed. When the agent decides to use experience, it emits a special trigger token, \texttt{[Retrieve]}, which activates retrieval. The full prompt template is provided in Appendix~\ref{app:prompt_template}.

Formally, at each step $t$, the agent first produces an intermediate reasoning trace:
\[
h_t = \pi_\theta(c_t),
\]
which reflects its current understanding of the state. We then check whether $h_t$ contains the trigger token. If the trigger appears, the current context is used as a retrieval query:
\[
M(c_t) = \text{Retrieve}(M, c_t),
\]
and the retrieved experience is incorporated for subsequent action generation:
\[
a_t = \pi_\theta([c_t; h_t; M(c_t)]).
\]
If the trigger token does not appear, the agent proceeds without additional experience incorporated. Overall, this yields the following decision loop:
\[
\texttt{reasoning} \rightarrow \texttt{(optional experience utilization)} \rightarrow \texttt{action}.
\]
\paragraph{Role of ExpWeaver.}

ExpWeaver should be viewed as a minimal instantiation for studying runtime experience utilization, rather than as a modification to experience construction. It introduces three useful properties for analysis: \textbf{(1) Adaptive usage.} Experience is invoked during reasoning only when the agent determines that additional guidance may be useful. \textbf{(2) Orthogonality.} ExpWeaver is independent of how experience is represented or updated, making it applicable to diverse self-evolving frameworks. \textbf{(3) Observability.} Because retrieval is triggered explicitly within the reasoning process, we can inspect when experience is used and analyze whether such usage is meaningful.

\subsection{Agentic Reinforcement Learning-based ExpWeaver}

\label{subsec:ExpWeaver_rl}

The prompting-based setup tests whether existing agents can be guided to regulate experience usage at inference time. We further ask whether this behavior can be strengthened through learning. To this end, we integrate ExpWeaver into experience-driven agentic reinforcement learning.

% \paragraph{Experience Usage as a Learnable Decision.}

% Under ExpWeaver, experience utilization can be viewed as a binary decision made during reasoning:
% \[
% u_t \in \{0, 1\},
% \]
% where $u_t = 1$ indicates that the agent invokes experience retrieval at step $t$. The agent policy therefore governs both action generation and experience usage:
% \[
% a_t, u_t \sim \pi_\theta(\cdot \mid c_t).
% \]
% This formulation allows reinforcement learning to optimize not only what action to take, but also when experience should be used.

\paragraph{Policy Optimization with GRPO.}

We train the policy using Group Relative Policy Optimization (GRPO) \citep{shao2024deepseekmath}. For an input $x$, the agent samples $G$ trajectories $\{\tau^{(1)}, \dots, \tau^{(G)}\}$. Each trajectory receives a binary reward $R_i = r(\tau^{(i)}) \in \{0,1\}$ indicating whether the task is successfully completed. The policy is updated with a PPO-style \citep{schulman2017proximal} clipped objective:
\[
\mathcal{J}(\theta) = \mathbb{E} \Bigg[ \frac{1}{G} \sum_{i=1}^{G} \min \big( r_i A_i, \text{clip}(r_i, 1-\epsilon, 1+\epsilon) A_i \big) - \beta D_{\text{KL}}(\pi_\theta \| \pi_{\text{ref}}) \Bigg],
\]
where $A_i$ is computed from intra-group normalized rewards, and the KL penalty keeps the updated policy close to the reference policy.

% \paragraph{Learning to Regulate Experience Use.}

% Since the generated trajectories include both actions and experience-trigger decisions, policy optimization reinforces usage patterns that improve task success. This allows us to examine whether coupling experience utilization with decision-making is merely a prompting-time effect, or whether it can become a learnable capability. In this sense, RL serves as a tool for testing whether agents can further learn how and when to use experience, rather than relying on fixed usage strategies.

\section{How Should Experience Be Utilized in Self-Evolving Agents?}

\subsection{Experimental Setup}

\noindent{\textbf{Experience Utilization Strategies.}}
We compare four experience utilization strategies. 
\textbf{w/o Experience} removes the experience repository and performs task execution without using any past experience. 
\textbf{Initialization-only (Init-only)} retrieves relevant experience once at the beginning of task execution and keeps it as static context throughout the trajectory. 
\textbf{Always-on} retrieves and injects experience at every decision step, regardless of whether it is needed for the current state. 
\textbf{ExpWeaver} interweaves experience utilization with the decision-making process, allowing the agent to invoke experience during reasoning only when additional guidance is needed.

\noindent{\textbf{Agent Framework.}}
We implement the four experience utilization strategies described above across four representative self-evolving frameworks. These frameworks cover diverse forms of experience representation: \textbf{ReasoningBank} \citep{ouyang2025reasoningbank} uses distilled insights, \textbf{AWM} \citep{wang2025agent} represents experience as executable workflows, \textbf{SkillRL} \citep{xia2026skillrl} models experience as reusable skills, and \textbf{G-Memory} \citep{zhang2025g} combines raw trajectories with distilled insights. G-Memory operates in a \emph{multi-agent} setting, while the others are \emph{single-agent} frameworks.

The vanilla versions of ReasoningBank, SkillRL, and G-Memory follow an Init-only usage strategy. AWM does not include an explicit retrieval mechanism; therefore, its Init-only and Always-on variants are equivalent in practice. These frameworks also cover both \emph{offline} and \emph{online} self-evolving paradigms: SkillRL relies on a pre-constructed experience repository, whereas the other frameworks dynamically accumulate experience through interaction with the environment.

This selection spans a broad spectrum of experience representations, agent settings, and evolution paradigms, enabling a comprehensive evaluation of how different experience utilization strategies affect self-evolving agents. Further details are provided in Appendix~\ref{app:agent_intro}.

\noindent{\textbf{Backbone Model.}}
We conduct experiments across a diverse set of seven LLMs from multiple families, including \textbf{GPT-5.2} \citep{singh2025openai}, \textbf{DeepSeek-V4-Pro} \citep{deepseekai2026deepseekv4}, and \textbf{Kimi-K2.5} \citep{kimiteam2026kimik25visualagentic}, as well as open-weight Qwen series including \textbf{Qwen3.5-397B-A17B} \citep{qwen3.5} and several \textbf{Qwen3} variants (32B, 14B and 4B) \citep{yang2025qwen3}, covering a wide range of model scales and architectures for comprehensive evaluation. Due to computational constraints, we conduct reinforcement learning experiments only on Qwen3-14B and Qwen3-4B.

\noindent{\textbf{Environment \& Benchmark.}}
We evaluate across 8 benchmarks in 3 environments:
(1) For \emph{embodied interaction}, we adopt the interactive environment \textbf{ALFWorld} \citep{shridhar2021alfworld}.
(2) For \emph{web navigation}, we evaluate on \textbf{WebShop} \citep{yao2022webshop}.
(3) For \emph{knowledge-intensive question answering (QA)}, we include \textbf{HotpotQA} \citep{yang2018hotpotqa}, \textbf{NQ} \citep{kwiatkowski2019natural} \textbf{TriviaQA} \citep{joshi2017triviaqa}, \textbf{2Wiki} \citep{ho2020constructing}, \textbf{MuSiQue} \citep{trivedi2022musique} and \textbf{Bamboogle} \citep{press2023measuring}.
These benchmarks span diverse environments and task types, providing a comprehensive evaluation of experience-driven adaptation (details in Appendix~\ref{app:env_bench}).

\noindent{\textbf{Implementation Details.}} For prompting-based experiments, we access those models via official APIs. All agentic reinforcement learning experiments are conducted on a cluster of eight NVIDIA A100 80GB GPUs.
Further implementation details are provided in Appendix~\ref{app:implement}.

\begin{figure}[t]
  \centering
  \includegraphics[width=\linewidth]{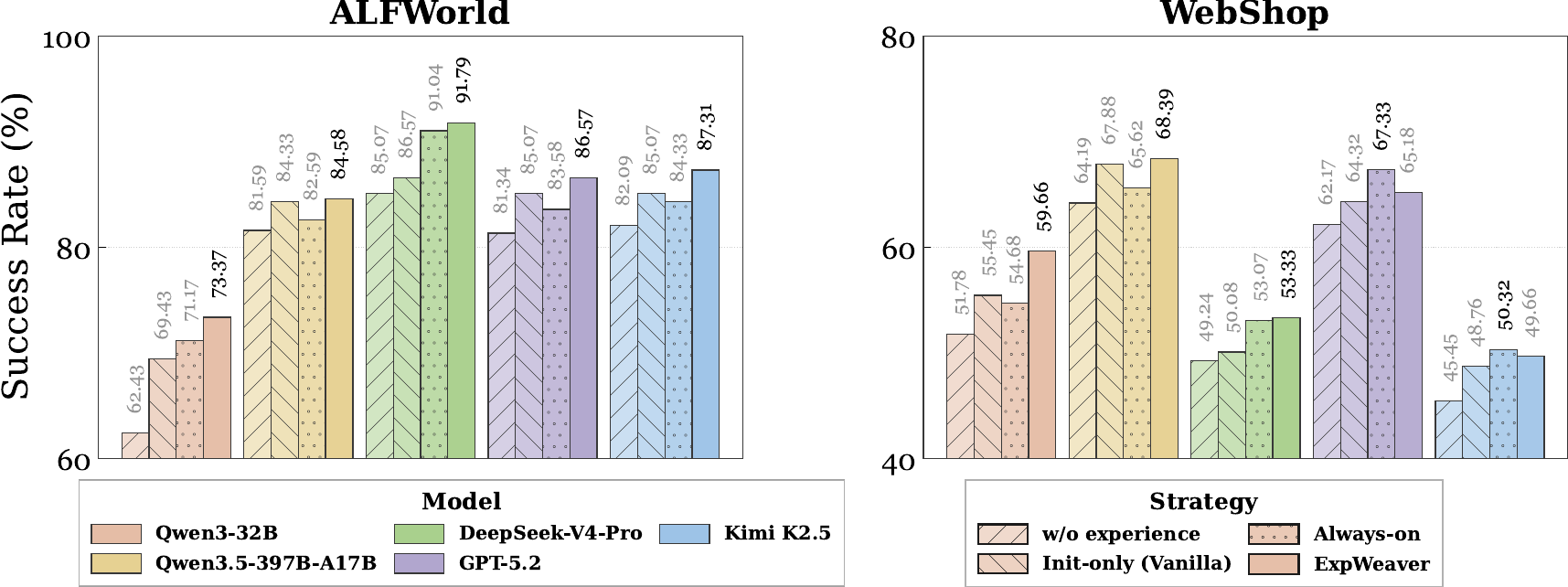}
  \caption{Prompting-based results under the ReasoningBank framework on ALFWorld and WebShop.}
  \label{fig:reasoningbank_main}
\end{figure}
\vspace{-1pt}
\begin{figure}[t]
  \centering
  \includegraphics[width=\linewidth]{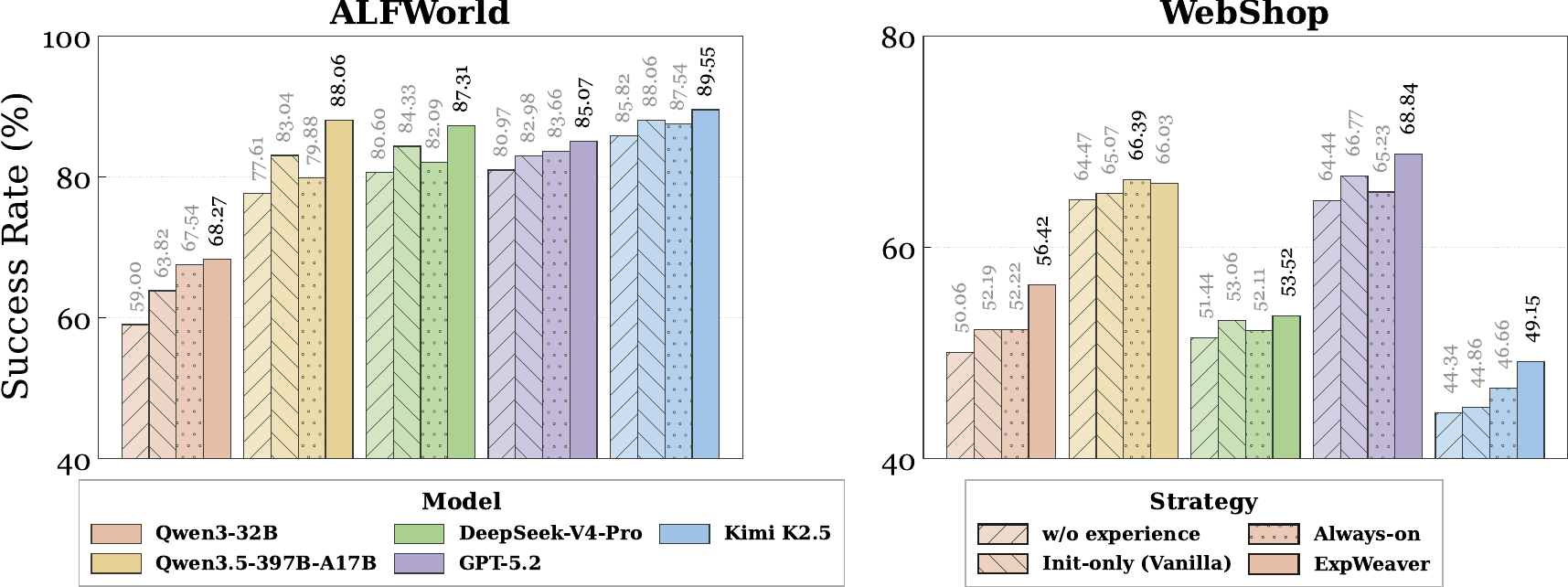}
  \caption{Prompting-based results under the SkillRL framework on ALFWorld and WebShop.}
  \label{fig:skillrl_main}
\end{figure}

\subsection{Results on Prompting-based ExpWeaver}
\label{subsec:prompting_results}

Figures~\ref{fig:reasoningbank_main} and~\ref{fig:skillrl_main} present the prompting-based results on ALFWorld and WebShop across ReasoningBank and SkillRL frameworks. Due to space constraints, results on knowledge-intensive QA tasks and the full results for G-Memory and AWM are provided in Appendix~\ref{app:prompting_results}. Overall, ExpWeaver consistently achieves the best performance among different experience utilization strategies across models, environments, and experience representations. This demonstrates the efficacy of interweaving experience utilization into the decision-making process. We highlight the following key observations:

\noindent{\textbf{Consistent gains over rigid experience usage strategies.}}
ExpWeaver consistently improves over both \emph{Init-only} and \emph{Always-on} baselines across nearly all settings. This trend holds across different backbones and environments. The results indicate that neither under-utilization nor over-utilization of experience is optimal; instead, integrating experience utilization into the decision-making process leads to more reasonable behavior and effective performance.

\noindent{\textbf{Robust effectiveness across diverse experience representations.}}
ExpWeaver achieves consistent improvements across frameworks with fundamentally different experience forms, including distilled insights (ReasoningBank), workflows (AWM), skills (SkillRL), and hybrid memory (G-Memory). This demonstrates that the benefits of interweaving experience utilization into the decision-making process are largely independent of how experience is represented, supporting the claim that experience {usage}, rather than representation, is the key bottleneck.

\noindent{\textbf{Strong generalization across agent settings and self-evolving paradigms.}}
ExpWeaver generalizes across both single-agent and multi-agent settings (i.e., G-Memory), as well as offline and online self-evolving paradigms. This is notable given their distinct experience construction and update mechanisms. In particular, SkillRL provides a controlled offline setting where ExpWeaver uses the same pre-constructed experience repository as the vanilla baseline and only changes runtime utilization. The gains in this setting indicate that the improvement comes from regulating experience usage itself, supporting ExpWeaver as a general mechanism across heterogeneous agent designs.

\begin{wrapfigure}{r}{0.4\textwidth}
  \vspace{-3.5mm}
  \centering
  \includegraphics[width=1.0\linewidth]{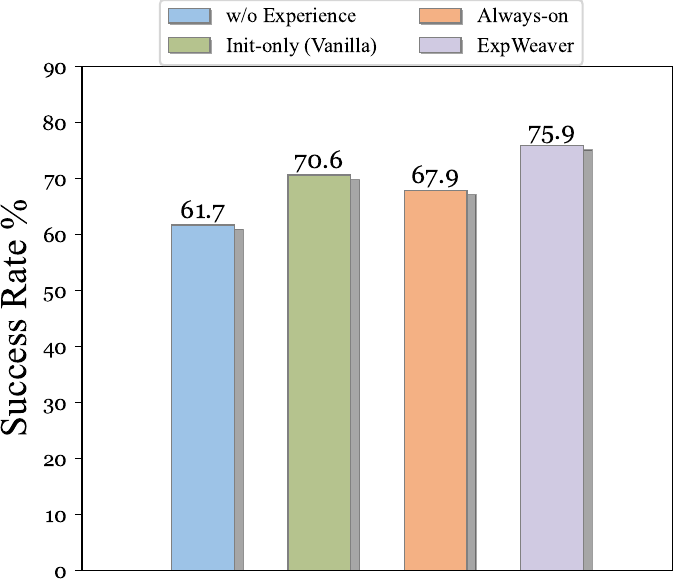}
  \vspace{-2mm}
  \caption{RL results of Qwen3-4B on ALFWorld after GRPO training.}
  \label{fig:rl_alfworld_4B}
  \vspace{-4mm}
\end{wrapfigure}

\subsection{Results on RL-based ExpWeaver}
\label{subsec:rl_results}

We further investigate whether interweaving experience utilization with the decision-making process can be strengthened through agentic reinforcement learning. Specifically, we adopt the ReasoningBank framework and train Qwen3-4B on the ALFWorld environment with GRPO under four experience utilization strategies. Detailed training settings are provided in Appendix~\ref{app:rl_details}.

Figure~\ref{fig:rl_alfworld_4B} shows that the way experience is used substantially affects RL performance. All experience-based variants outperform w/o Experience, confirming the value of experience under GRPO training. Yet Always-on underperforms Init-only, suggesting that excessive experience exposure can introduce interference. ExpWeaver achieves the highest success rate, improving over Init-only by 5.3 points and Always-on by 8.0 points. This indicates that RL benefits most when experience utilization is interwoven with decision-making rather than fixed as static context or injected at every step.

We also observe consistent trends on knowledge-intensive QA with Qwen3-14B. As shown in Appendix~\ref{app:rl_qa}, ExpWeaver generally achieves stronger learning dynamics and higher final success rates than the vanilla baseline across QA benchmarks such as NQ, TriviaQA, Bamboogle, and 2Wiki. These results suggest that the learnability of experience utilization is not limited to embodied interaction, but also generalizes to reasoning-intensive QA scenarios.

\section{Analysis and Discussions}

\begin{table*}[t]
\centering
\caption{
Average number of experience retrievals per sample under ExpWeaver. 
We report results on ReasoningBank and G-Memory.
}
\label{tab:pattern}
\resizebox{\linewidth}{!}{
\begin{tabular}{lcccccc}
\toprule
\multirow{2.5}{*}{\textbf{Model}}
& \multicolumn{3}{c}{\textbf{ReasoningBank}}
& \multicolumn{3}{c}{\textbf{G-Memory}} \\

\cmidrule(lr){2-4} \cmidrule(lr){5-7}

& ALFWorld & WebShop & HotpotQA & ALFWorld & WebShop & HotpotQA  \\ 
\midrule
\texttt{Qwen3-32B} &2.17 &0.13 &0.42 &1.87 &0.04 &0.99 \\
\texttt{Qwen3.5-397B-A17B} &0.43 &0.00 &0.00 &0.75 &0.00 &0.01 \\
\texttt{DeepSeek-V4-Pro} &0.82 &0.95 &0.17 &0.94 &0.16 &0.18 \\
\texttt{GPT-5.2} &0.00 &0.00 &0.00 &0.00 &0.15 &0.01 \\
\texttt{Kimi-K2.5} &1.14 &0.02 &0.03 &2.31 &0.00 &0.18 \\
\bottomrule
\end{tabular}
}
\end{table*}

\subsection{Usage Pattern Analysis of ExpWeaver}
\label{subsec:pattern_analysis}

To understand how ExpWeaver utilizes experience in practice, we analyze its usage patterns from two perspectives: (1) a \emph{horizontal comparison} across tasks, measured by the average number of experience retrievals per sample; and (2) a \emph{temporal analysis} within trajectories, measured by the fraction of trajectories that invoke experience at each step.

\noindent{\textbf{Horizontal comparison across tasks.}}
Table~\ref{tab:pattern} reports the average number of experience retrievals per sample under two representative frameworks, ReasoningBank and G-Memory. Additional results on AWM and SkillRL are provided in Appendix~\ref{app:usage_pattern}. Overall, ExpWeaver exhibits a clear task-aware, model-aware, and framework-agnostic usage pattern.

First, experience usage is highly task-dependent. Across most frameworks and backbones, ALFWorld induces substantially more retrievals than WebShop and HotpotQA, suggesting that experience is more frequently needed in embodied interaction, where agents must perform long-horizon planning, track partial observations, and recover from uncertain states. Second, retrieval frequency tends to decrease as model capability increases. Stronger backbone models such as GPT-5.2 and Qwen3.5-397B-A17B rarely invoke experience, while smaller models such as Qwen3-32B rely on it more often. This indicates that ExpWeaver does not mechanically retrieve experience, but treats it as a complementary resource when the model requires additional guidance. Third, similar patterns appear across different self-evolving frameworks, despite their distinct experience representations. This suggests that the observed behavior is not tied to a specific memory format, but is induced by interweaving experience utilization into the decision-making process.

\begin{wrapfigure}{r}{0.5\textwidth}
  \vspace{-3.5mm}
  \centering
  \includegraphics[width=1.0\linewidth]{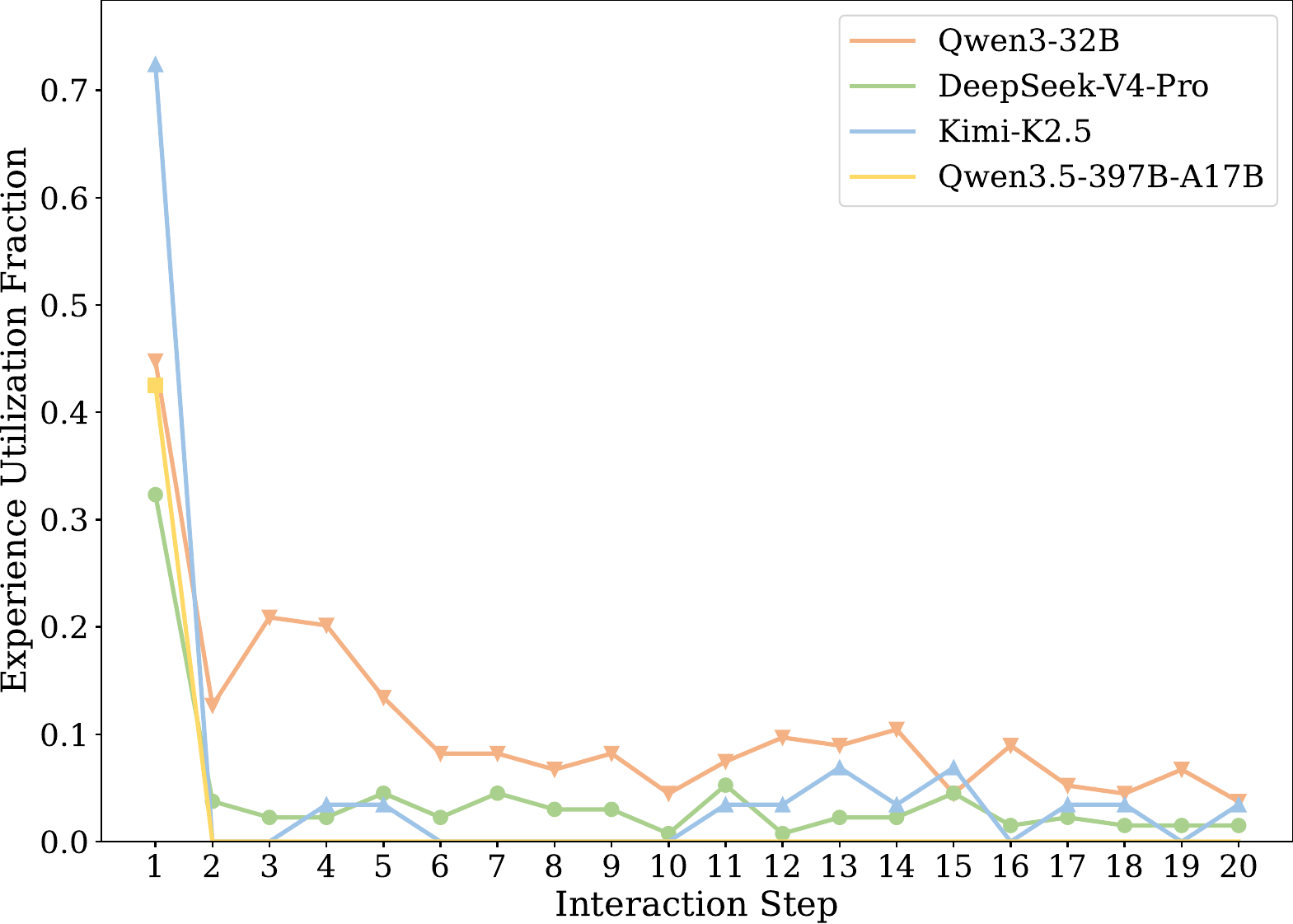}
  \vspace{-2mm}
  \caption{Temporal pattern of experience utilization under ExpWeaver on ALFWorld with the ReasoningBank framework.}
  \label{fig:temporal_rb_alfworld}
  \vspace{-4mm}
\end{wrapfigure}

\noindent{\textbf{Temporal analysis within decision trajectories.}}
Figure~\ref{fig:temporal_rb_alfworld} reports the fraction of trajectories that invoke experience at each interaction step in ALFWorld under the ReasoningBank framework. Experience utilization shows a clearly non-uniform temporal pattern: retrieval is concentrated at the first step, drops sharply afterward, and occasionally resurges at later stages. This suggests that agents mainly consult experience when task understanding and high-level planning are most critical, rather than using it uniformly throughout the trajectory.

Importantly, later retrievals are not random. In ALFWorld, they often occur during object-search phases, where the agent must locate target items under partial observability. When search attempts fail repeatedly, experience is re-invoked to provide corrective guidance. This suggests that experience utilization may be associated with moments of increased decision uncertainty, a hypothesis we further examine through entropy-based analysis in \S\ref{subsec:entropy_analysis}. Additional temporal analyses across more frameworks and tasks are provided in Appendix~\ref{app:temporal_analysis}, and specific qualitative examples of ExpWeaver on different tasks are also provided in Appendix~\ref{app:qualitative_examples}.

\begin{table*}[t]
\centering
\caption{Causal ablation of ExpWeaver-invoked experience utilization. 
{Empty} removes retrieved experience at positions where ExpWeaver chooses to invoke it, while {Random Utilization} invokes experience at randomly selected positions.}
\label{tab:ablation}
\resizebox{\linewidth}{!}{
\begin{tabular}{lcccccc}
\toprule
\multirow{2.5}{*}{\textbf{Method}}
& \multicolumn{3}{c}{\textbf{ReasoningBank}}
& \multicolumn{3}{c}{\textbf{G-Memory}} \\

\cmidrule(lr){2-4} \cmidrule(lr){5-7}

& ALFWorld & WebShop & HotpotQA & ALFWorld & WebShop & HotpotQA  \\ 
\midrule
\rowcolor{gray!8} \multicolumn{7}{c}{\texttt{Qwen3-32B}} \\
\midrule
ExpWeaver &\textbf{73.37} &\textbf{59.66} &\textbf{45.33} &\textbf{83.33} &\textbf{45.56} &\textbf{47.00} \\
Random Utilization &68.46 &49.94 &44.67 &71.64 &37.72 &45.33 \\
Empty &71.64 &54.56 &43.00 &75.37 &39.53 &45.33 \\
\midrule
\rowcolor{gray!8} \multicolumn{7}{c}{\texttt{DeepSeek-V4-Pro}} \\
\midrule
ExpWeaver &\textbf{91.79} &\textbf{53.33} &\textbf{55.00} &\textbf{88.81} &\textbf{66.02} &\textbf{55.00} \\
Random Utilization &84.33 &48.29 &52.00 &81.34 &62.11 &52.67 \\
Empty &86.57 &50.92 &53.00 &85.82 &64.09 &53.67 \\
\bottomrule
\end{tabular}
}
\end{table*}

\subsection{Causal Effect of ExpWeaver-Invoked Experience Utilization}
\label{subsec:causal_ablation}

We further examine whether the observed ExpWeaver's non-uniform usage pattern is causally beneficial. We compare ExpWeaver with two ablations: \textbf{Empty}, which removes retrieved experience at positions invoked by ExpWeaver, and \textbf{Random Utilization}, which invokes experience at random positions. For Random Utilization, we first compute the average number of experience invocations and the average trajectory length of ExpWeaver on each benchmark, and then invoke experience at each step with probability given by their ratio.

As shown in Table~\ref{tab:ablation}, both ablations consistently underperform ExpWeaver. We focus on Qwen3-32B and DeepSeek-V4-Pro because these two backbones invoke experience more frequently under ExpWeaver, providing enough usage events for reliable ablation analysis. For Qwen3-32B, removing experience leads to clear drops across ReasoningBank and G-Memory, while random utilization also fails to match ExpWeaver. DeepSeek-V4-Pro exhibits the same trend, with ExpWeaver consistently outperforming both ablation variants across all evaluated settings.

These results provide causal evidence that interweaving experience utilization with decision-making is effective because the agent learns \emph{when} experience should enter the reasoning process.

\subsection{Mechanistic Analysis: Experience Usage under Uncertainty}
\label{subsec:entropy_analysis}

\begin{wrapfigure}{r}{0.5\textwidth}
  \vspace{-3.5mm}
  \centering
  \includegraphics[width=1.0\linewidth]{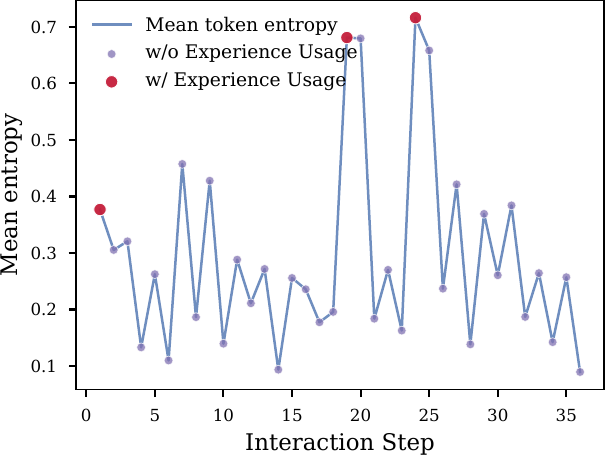}
  \vspace{-2mm}
  \caption{Mechanistic analysis of experience usage in ExpWeaver. 
The blue curve shows the mean token entropy at each reasoning step, while red points denote steps where experience is invoked.
}
  \label{fig:entropy}
  \vspace{-4mm}
\end{wrapfigure}

We further analyze what happens inside the agent when ExpWeaver decides to invoke experience. Our hypothesis is that experience usage is triggered when the agent encounters higher decision uncertainty. To examine this, we measure the token-level entropy of the agent's output distribution during each reasoning step, and compare the entropy at positions with and without experience usage.

Figure~\ref{fig:entropy} presents the entropy analysis results on ALFWorld. The blue curve denotes the mean token entropy at each step, while red points indicate steps where ExpWeaver invokes experience. We observe that experience usage tends to occur at entropy peaks. In contrast, steps without experience usage generally correspond to lower-entropy regions. This suggests that ExpWeaver is more likely to retrieve experience when the agent is uncertain about its next decision.

\section{Conclusion}

This work studies how experience should be utilized in self-evolving agents. We show that existing rigid usage schemes overlook an important design dimension: when and how experience should enter decision-making.
To examine this question, we introduce ExpWeaver, a lightweight instantiation that keeps experience construction unchanged while modifying runtime utilization. Across diverse frameworks, models, and environments, ExpWeaver consistently outperforms rigid usage strategies. Further analyses show that agents invoke experience selectively, at causally beneficial positions, and under higher reasoning uncertainty.
Overall, our findings suggest that effective self-evolution depends not only on {what} experience is stored, but also on {how} and {when} it is used.

\bibliography{natbib}
\bibliographystyle{unsrtnat}

%%%%%%%%%%%%%%%%%%%%%%%%%%%%%%%%%%%%%%%%%%%%%%%%%%%%%%%%%%%%

\appendix

\section{Limitations and Impact Statement}
\label{app:limitations}

\paragraph{Limitations.}
This work studies how experience should be utilized in self-evolving agents and shows that interweaving experience utilization with decision-making can improve performance across diverse frameworks and environments. Nevertheless, several limitations remain.

First, ExpWeaver is implemented through prompting and trigger-based retrieval, which makes it lightweight and broadly applicable, but also means that its behavior depends on the instruction-following ability of the underlying LLM. We observe consistent gains across multiple backbones, but weaker small-scale models may fail to reliably decide when experience is needed through prompting.

Second, our analysis focuses on experience utilization under existing experience construction pipelines. While this design helps isolate the effect of runtime utilization, it does not jointly optimize experience construction and utilization. A promising direction is the recent trend of fully automated experience memory design, where agents can automatically discover, refine, and reorganize experience representations over time \citep{xiong2026learning,pan2026m}. Combining such automated memory construction with adaptive utilization may provide a more unified path toward self-evolving agents that can improve not only {what} they store, but also {how} and {when} they use it.

Finally, our current study primarily evaluates task performance. Other dimensions, such as safety, robustness, calibration, and long-term stability under continuous deployment, remain important directions for future investigation \citep{shao2025your,zhao2026safety,zhao2026large}.

\paragraph{Impact Statement.}
This work highlights experience utilization as a critical design dimension for self-evolving agents. By showing that agents benefit from selectively invoking experience during decision-making, our findings may help build more adaptive, efficient, and interpretable agent systems. In particular, ExpWeaver provides a lightweight way to improve existing self-evolving frameworks without modifying their experience construction pipelines, which may lower the cost of deploying experience-driven agents in diverse environments.

At the same time, more effective experience utilization may also amplify both desirable and undesirable behaviors contained in accumulated experience. If the experience repository contains biased, unsafe, outdated, or task-inappropriate content, selectively invoking such experience may still mislead the agent. Therefore, practical deployment should be accompanied by careful experience filtering, monitoring, and safety evaluation. We hope this work encourages future research to study not only how agents accumulate experience, but also how experience should be governed, audited, and safely integrated into decision-making.

\section{Self-Evolving Agents}
\label{app:agent_intro}

We present detailed description of the four experience-driven self-evolving agents used in our experiments: ReasoningBank \citep{ouyang2025reasoningbank}, AWM\citep{wang2025agent}, SkillRL\citep{xia2026skillrl} and G-Memory \citep{zhang2025g}. Across all frameworks, agents adapt their behavior by accumulating, retrieving, and reusing external experiences stored in explicit memory structures, rather than by updating the parameters of the underlying language model. In the following, we summarize the core design principles and memory mechanisms of these agents in detail:

\begin{itemize}[leftmargin=*]
    \item \textbf{ReasoningBank} \citep{ouyang2025reasoningbank} embodies an instance of an online self-evolving agent. It keeps an ever-expanding memory that records condensed reasoning patterns distilled from the agent's past interactions, covering both successful cases and failures. Following each task completion, the agent assesses its own performance and selectively incorporates newly acquired experiences into this memory repository. During inference, pertinent reasoning strategies are retrieved from the memory and incorporated into the agent's context to guide future interactions. The system forms a closed feedback loop where experiences are continuously accumulated, accessed, and reused throughout deployment, enabling the agent's behavior to progressively adapt over time.
    \item \textbf{AWM} \citep{wang2025agent} embodies another instance of an online self-evolving agent. It maintains a growing memory of reusable workflows—common sub‑routines distilled from successful agent trajectories, with example‑specific values abstracted into variables. After each task, an LLM‑based evaluator judges whether the trajectory succeeded; if so, the agent induces one or more workflows from the trajectory and adds them to the memory. During inference, relevant workflows are retrieved and incorporated into the agent’s context to guide future actions. AWM operates both offline (inducing workflows from annotated training examples) and online (learning purely from test queries in a streaming fashion). This closed‑loop process enables the agent to accumulate increasingly complex workflows over time, rapidly improving performance from only a handful of examples and generalising across tasks, websites, and domains.
    \item \textbf{SkillRL} \citep{xia2026skillrl} is an online self-evolving agent that bridges raw experience and policy improvement via automatic skill discovery and recursive evolution. It maintains a hierarchical skill library—SKILLBANK—comprising general strategies and task‑specific heuristics, both distilled from successful and failed trajectories by a teacher model. During inference, relevant skills are retrieved to augment the agent’s context, achieving 10–20× token compression over raw trajectories. A cold‑start supervised fine‑tuning stage teaches the base model how to use skills, followed by reinforcement learning (GRPO) that jointly optimises the policy and recursively evolves the skill library: failed validation trajectories trigger generation of new skills, enabling the library and policy to co‑adapt. This closed‑loop process yields faster convergence, higher asymptotic performance, and strong generalisation across embodied and web tasks.
    \item \textbf{G-Memory} \citep{zhang2025g} is an online self-evolving memory mechanism designed for multi-agent systems. It maintains a shared, persistent memory that records past multi-agent collaboration experiences across tasks, capturing both abstract insights and condensed interaction histories. When a new task arrives, relevant memory entries are retrieved and selectively injected into the contexts of different agents to support coordination and reasoning. After task completion, newly generated interactions and distilled insights are incorporated into the memory, updating its contents during deployment. This continual retrieval-and-update process enables agent teams to accumulate and reuse collaborative experience over time, allowing collective behavior to adapt without modifying the underlying language models.
\end{itemize}

\section{Environment and Benchmark}
\label{app:env_bench}

\subsection{Embodied Interaction}

\begin{itemize}[leftmargin=*]
    \item \textbf{ALFWorld} \citep{shridhar2021alfworld} is an embodied benchmark that aligns abstract, text-based environments with interactive visual-based scenes to execute household tasks. It builds on the ALFRED \citep{shridhar2020alfred} benchmark by providing paired representations of the same underlying tasks, where agents can operate through high-level textual commands in a simulated environment. The tasks span multiple categories such as pick-and-place, cleaning, heating, and cooling, and require multi-step interaction with objects and receptacles distributed across diverse room layouts. The benchmark is constructed to maintain a shared underlying world state across modalities, enabling consistent correspondence between language-level actions and embodied executions. We utilized the 134 solvable tasks.
\end{itemize}

\subsection{Web Navigation}

\begin{itemize}[leftmargin=*]
    \item \textbf{WebShop} \citep{yao2022webshop} is a large-scale web interaction benchmark that simulates realistic online shopping scenarios through a self-contained e-commerce environment. It includes over one million real-world products and 12,087 crowdsourced natural language instructions, each specifying a product requirement to be fulfilled through a sequence of web-based actions. Given an instruction, an agent must navigate search results, inspect product pages, select appropriate options, and complete a purchase to satisfy the specified constraints. We follow the instructions of ExpeL to set the implementation details of the WebShop Environment.
\end{itemize}

\subsection{Knowledge-Intensive Question Answering}

\begin{itemize}[leftmargin=*]
    \item \textbf{HotpotQA} \citep{yang2018hotpotqa} is a large-scale question answering benchmark designed to support multi-hop reasoning over natural language text. It contains question–answer pairs constructed from Wikipedia articles, where answering each question requires reasoning across multiple supporting documents. The questions are diverse in form, including standard factoid queries as well as comparison and yes/no questions, and are not constrained by predefined knowledge base schemas. 
    \item \textbf{TriviaQA} \citep{joshi2017triviaqa} is a challenging reading comprehension dataset containing over 650K question–answer–evidence triples. It consists of 95K question-answer pairs authored by trivia enthusiasts, with evidence documents provided for distant supervision. The questions have relatively complex, compositional syntax and often require cross‑sentence reasoning to answer.
    \item \textbf{2WikiMultiHopQA} \citep{ho2020constructing} is constructed using a hybrid of templated generation and logical rules based on Wikidata. Each question requires multi‑step reasoning across two Wikipedia articles, and the dataset includes explicit evidence information that reveals the full reasoning path from the question to the answer.
    \item \textbf{Natural Questions} (NQ) \citep{kwiatkowski2019natural} is derived from real, anonymized Google search queries. The dataset contains over 300,000 training examples, each annotated by human raters with both long and short answers from the corresponding Wikipedia page, providing a realistic benchmark for open‑domain QA.
    \item \textbf{Bamboogle} \citep{press2023measuring} is a small yet challenging hand‑crafted multi‑hop QA dataset designed to measure the compositional reasoning gap in language models. Each question explicitly requires combining information from two separate facts, and it is commonly used as a difficult out‑of‑domain evaluation set.

\end{itemize}
For each of the above datasets following the official settings of evaluated self-evolving frameworks, we randomly selected 100 questions from its development split for our experiments.

\section{Implementation Details}
\label{app:implement}

\subsection{Prompting-based Setup}
\label{app:prompt_template}

\begin{figure}[t] 
\centering
\includegraphics[width=\textwidth]{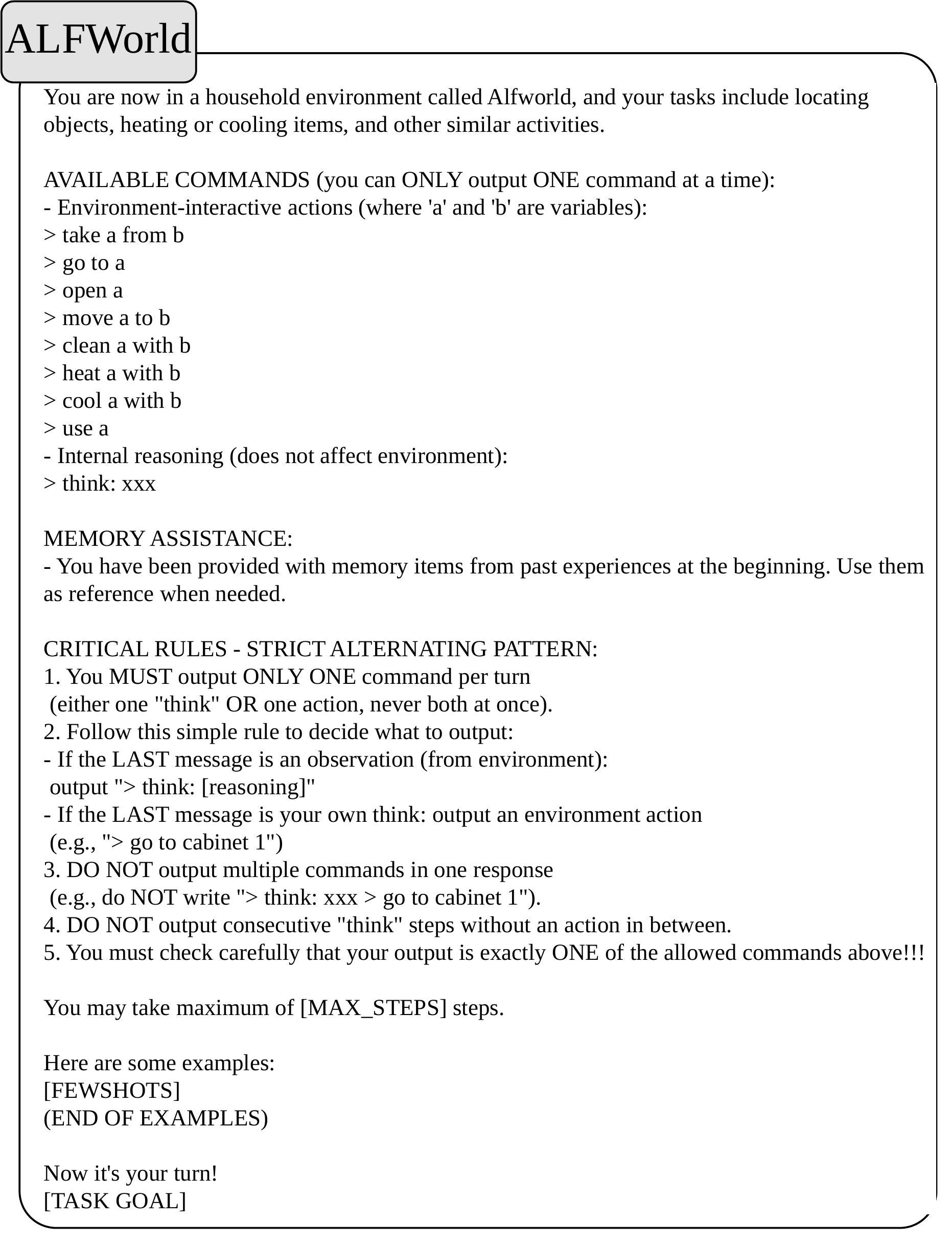}
\caption{System prompt on ALFWorld.}
\label{fig:prompt_alfworld}
\end{figure}

\begin{figure}[t] 
\centering
\includegraphics[width=\textwidth]{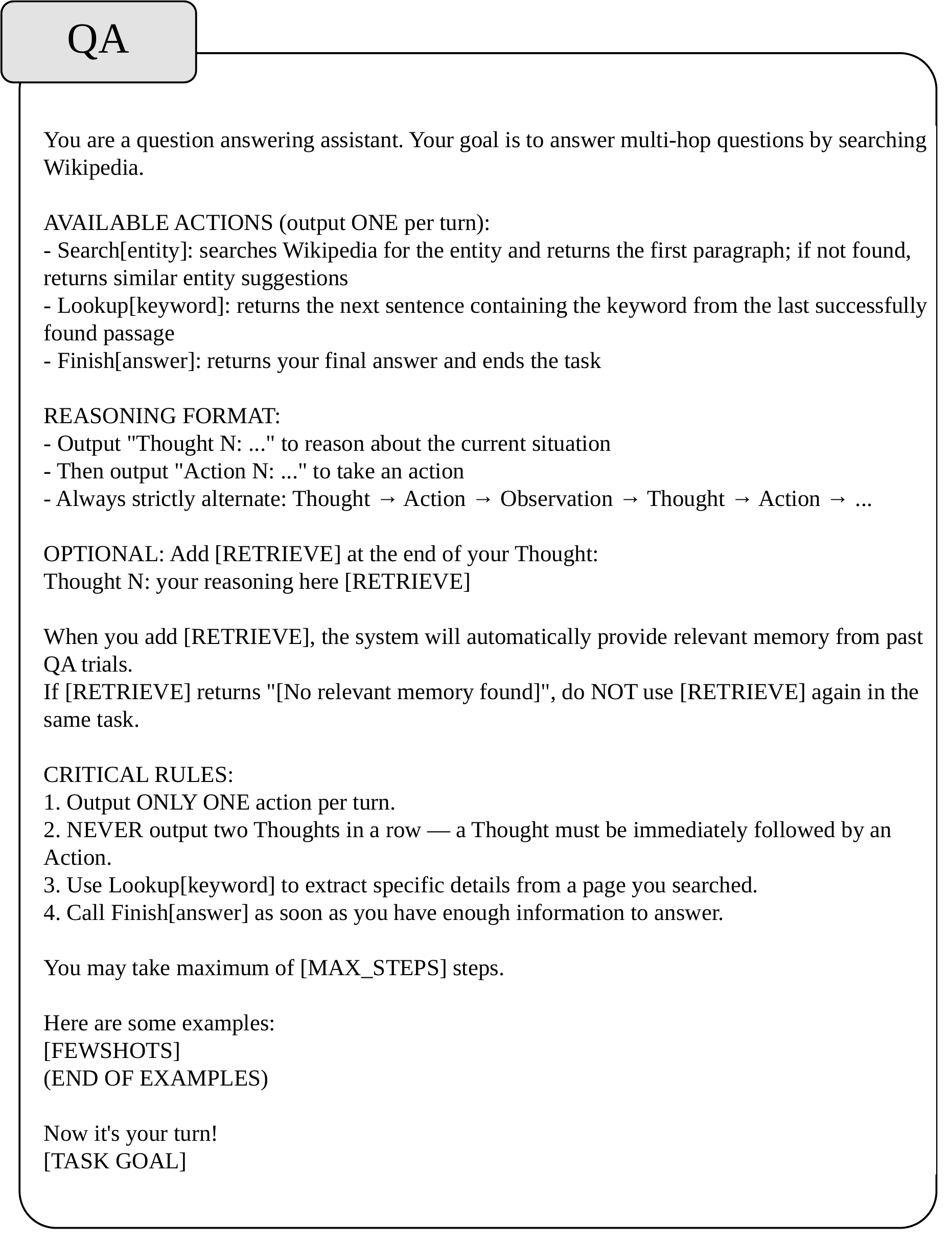}
\caption{System prompt on QA tasks.}
\label{fig:prompt_qa}
\end{figure}

\begin{figure}[t] 
\centering
\includegraphics[width=\textwidth]{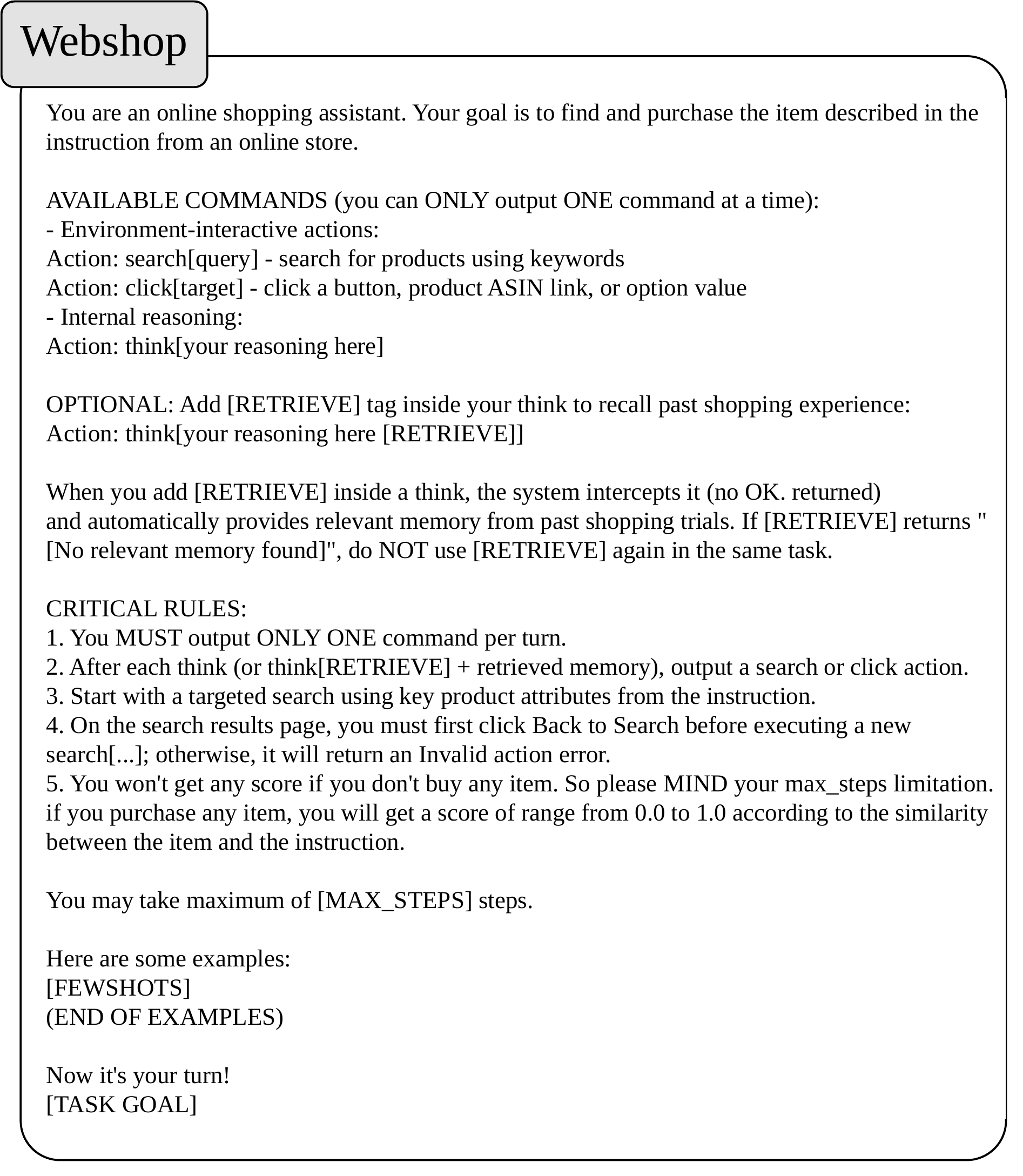}
\caption{System prompt of WebShop. }
\label{fig:prompt_webshop}
\end{figure}

ExpWeaver preserves the original design of each self-evolving agent framework with respect to experience construction, representation, and updating. Specifically, we leave the mechanisms for extracting, storing, and maintaining experience unchanged, and instead introduce only lightweight prompting. This is achieved by only augmenting the system prompt with instructions that interweave experience utilization with the reasoning process. We present the prompt designs for ALFWorld, QA tasks, and WebShop in Figure \ref{fig:prompt_alfworld}, Figure \ref{fig:prompt_qa} and Figure \ref{fig:prompt_webshop}, respectively. The detailed hyperparameter settings in our experiments for each self-evolving framework are described below. All reported results in our experiments are averaged over 3 independent runs.

\begin{itemize}[leftmargin=*]
    \item \textbf{ReasoningBank} Our experimental setup largely follows the configuration of ReasoningBank \citep{ouyang2025reasoningbank}. For the agent, we set the decoding temperature to 0.7 and adopt greedy decoding as the decoding strategy. On the WebArena benchmark, each task is allowed a maximum of 30 interaction steps. We use \emph{text-embedding-ada-002} as the embedding model to encode queries and memory items, and employ cosine similarity for retrieval. For each new query, the agent retrieves the top-{3} most relevant memory items, which are then injected into the agent's prompt. \
    \item \textbf{Agent Workflow Memory (AWM)} We adopt AWM \citep{wang2025agent} as a baseline. Following the original setup, we use temperature 0.0. Workflows are induced from successful trajectories, and during online inference the agent uses relevant workflows, which means no additional training data is used; the agent learns entirely from test queries in a streaming fashion.
    \item \textbf{G-Memory} We follow the experimental setup of G-Memory \citep{zhang2025g}. For coarse-grained retrieval over the query graph, queries are encoded using a MiniLM sentence embedding model and matched with cosine similarity, retrieving the top-\textit{k} most similar historical queries. For fine-grained retrieval, we further select the top-\textit{M} relevant queries using an LLM-based relevance scorer and sparsify their interaction graphs with an LLM-based graph compression module. The values of \textit{k} and \textit{M} are treated as tunable hyperparameters. Query nodes are labeled with execution status from {Failed, Resolved}. All retrieved insights and interaction subgraphs are injected into agent prompts before task execution, and the memory graphs are updated after each task without any gradient-based training.
    \item \textbf{SkillRL} We follow SkillRL \citep{xia2026skillrl} for inference. The agent uses a hierarchical SkillBank with general and task-specific skills. For each query, it retrieves top-\(6\) relevant task-specific skills (similarity threshold \(0.4\)) using an embedding model (\texttt{text-embedding-ada-002}) and includes all general skills. Retrieved skills are added to the prompt, and the agent decodes with temperature \(0.7\). The skill library is fixed during testing.
\end{itemize}

\subsection{Reinforcement Learning-based Setup}
\label{app:rl_details}

We conduct RL training on two types of interactive environments: ALFWorld and a suite of search‑augmented question answering (QA) tasks. For ALFWorld, we use the standard training splits with 3553 training tasks. All reported results are averaged over 3 independent runs. For QA training, we follow prior work and merge the training sets of NQ and HotpotQA to form a unified dataset for ExpWeaver. Out‑of‑domain generalization is assessed on TriviaQA, 2Wiki, and Bamboogle, where we sample 100 test instances per dataset. Hyperparameters used for ExpWeaver RL training are summarized in Table~\ref{tab:rl_hyperparams}. Our implementation is built on the verl-agent framework \citep{feng2025group}. Training runs for a fixed number of epochs or until validation performance converges.

\begin{table*}[t]
\centering
\caption{Hyperparameters for ExpWeaver training.}
\label{tab:rl_hyperparams}
\begin{tabular}{l c}
\toprule
\textbf{Hyperparameter} & \textbf{Value} \\
\midrule
\multicolumn{2}{c}{\textbf{RL Training}} \\
Learning rate & 1e-6 \\
Batch size & 8 \\
Group size \(G\) & 8 \\
Gradient accumulation steps & 64 \\
KL penalty coefficient \(\beta\) & 0.01 \\
Max prompt length (tokens) & 32768 \\
Max response length (tokens) & 512 \\
Total training epochs & 150 \\
\bottomrule
\end{tabular}
\end{table*}

\begin{figure}[t]
  \centering
  \includegraphics[width=\linewidth]{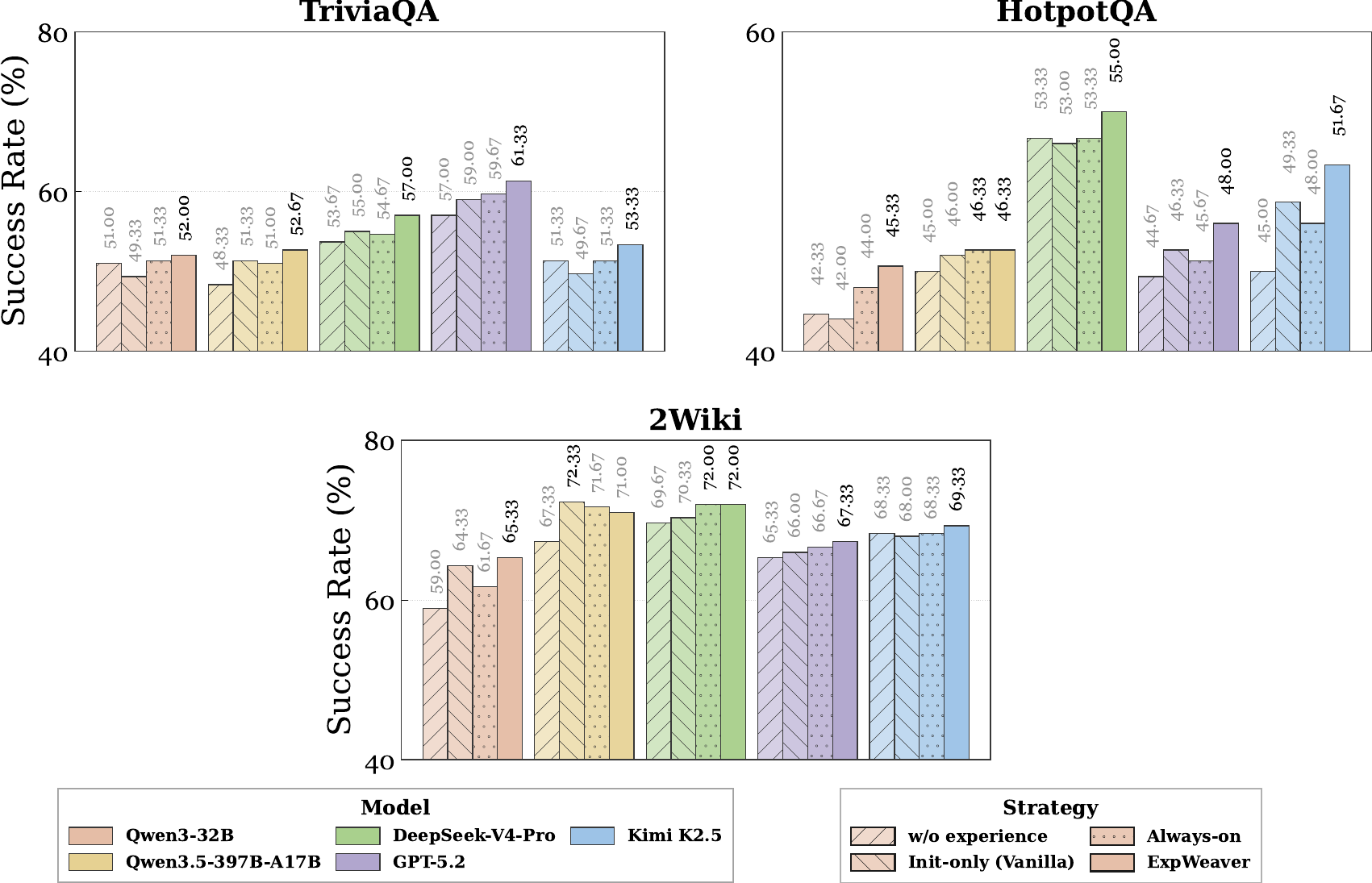}
  \caption{Prompting-based results under the ReasoningBank framework on three QA tasks.}
  \label{fig:reasoningbank_qa}
\end{figure}

\begin{figure}[t]
  \centering
  \includegraphics[width=\linewidth]{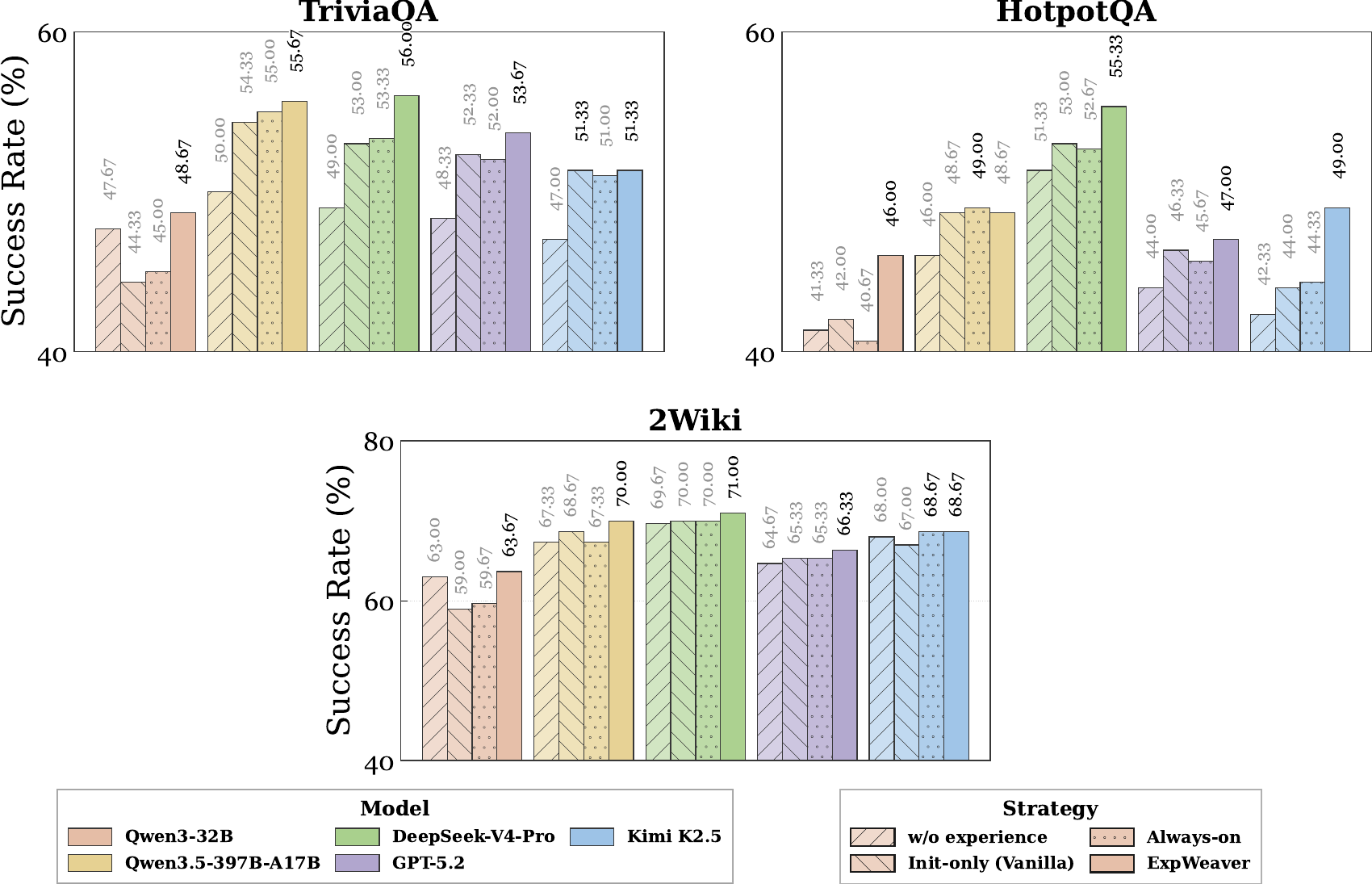}
  \caption{Prompting-based results under the SkillRL framework on three QA tasks.}
  \label{fig:skillrl_qa}
\end{figure}

\begin{figure}[t]
  \centering
  \includegraphics[width=\linewidth]{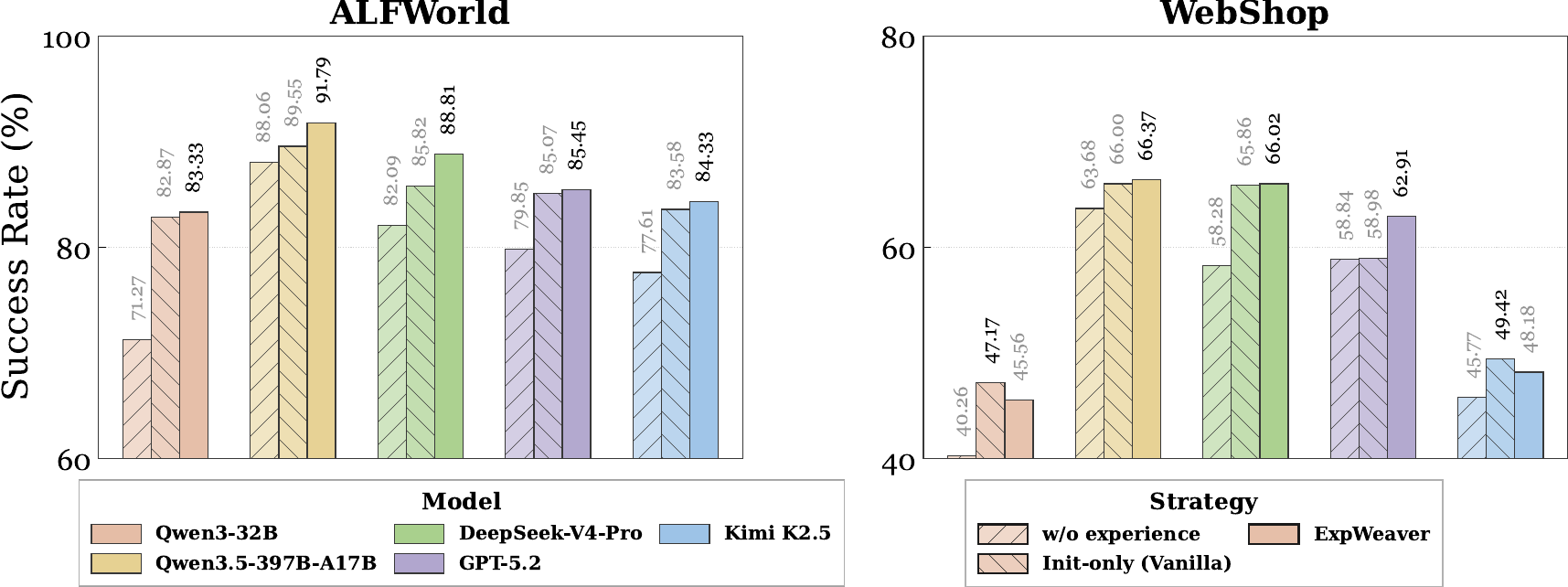}
  \caption{Prompting-based results under the G-Memory framework on ALFWorld and WebShop.}
  \label{fig:g_memory_app}
\end{figure}

\begin{figure}[t]
  \centering
  \includegraphics[width=\linewidth]{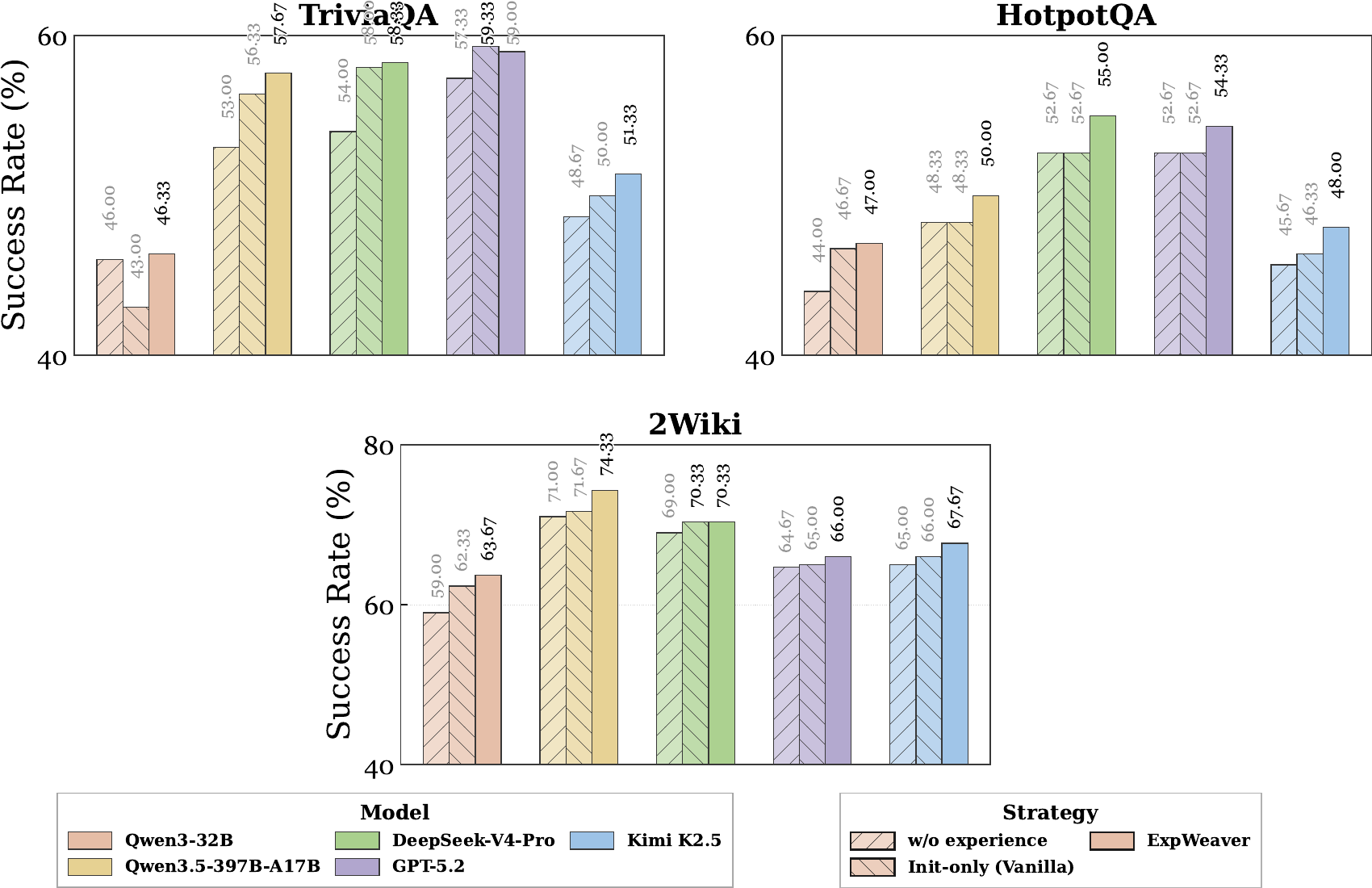}
  \caption{Prompting-based results under the G-Memory framework on three QA tasks.}
  \label{fig:g_memory_qa}
\end{figure}

\begin{figure}[t]
  \centering
  \includegraphics[width=\linewidth]{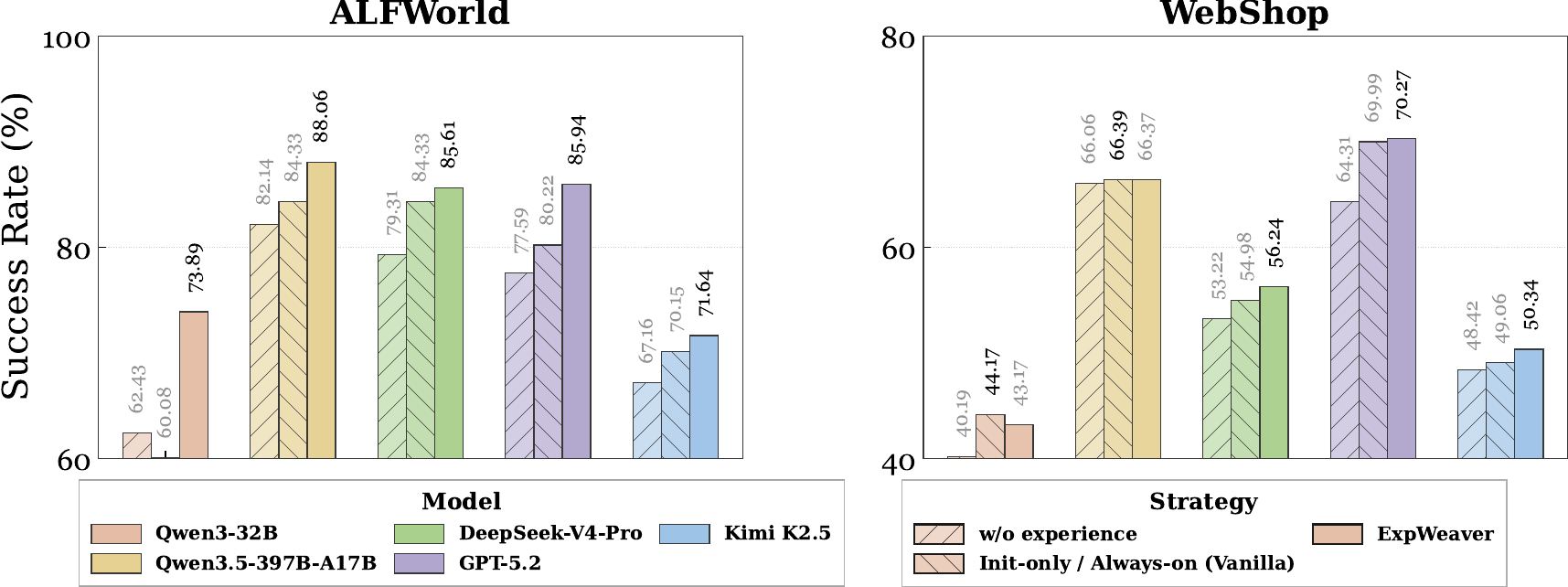}
  \caption{Prompting-based results under the AWM framework on ALFWorld and WebShop.}
  \label{fig:awm_app}
\end{figure}

\begin{figure}[t]
  \centering
  \includegraphics[width=\linewidth]{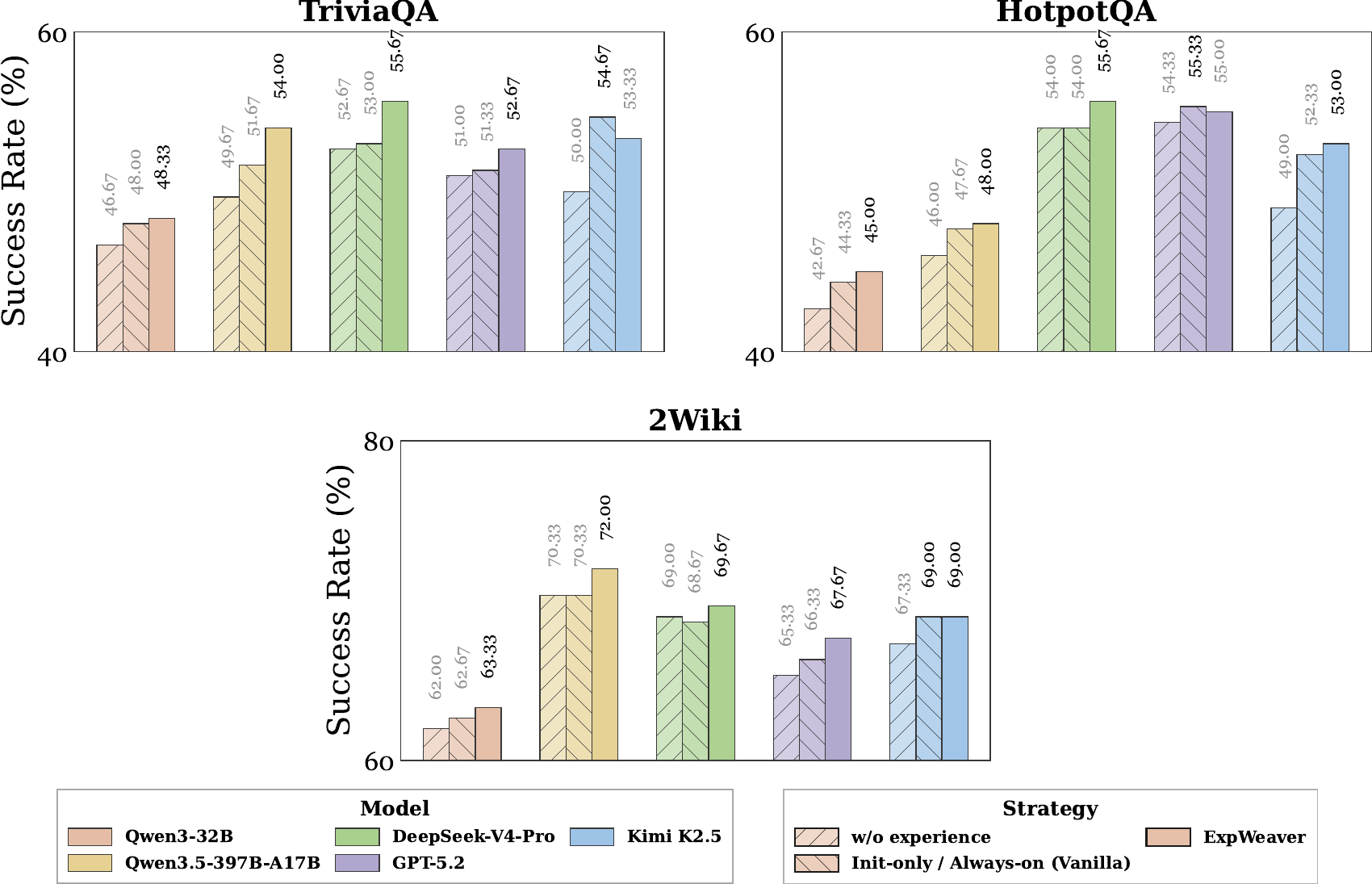}
  \caption{Prompting-based results under the AWM framework on three QA tasks.}
  \label{fig:awm_qa}
\end{figure}

\section{Additional Results with ExpWeaver}

\subsection{Additional Prompting-based Results}
\label{app:prompting_results}

\paragraph{Results on QA tasks.}
Figures~\ref{fig:reasoningbank_qa} and~\ref{fig:skillrl_qa} present the prompting-based results on three knowledge-intensive QA tasks under the ReasoningBank and SkillRL frameworks, respectively. Overall, ExpWeaver consistently achieves the best or competitive performance across TriviaQA, HotpotQA, and 2Wiki, further confirming the effectiveness of interweaving experience utilization into the decision-making process.

Under ReasoningBank, ExpWeaver improves over the vanilla initialization-only strategy in most settings. The gains are particularly clear on TriviaQA and HotpotQA, where ExpWeaver achieves the best performance across nearly all backbone models. On 2Wiki, ExpWeaver also generally improves over rigid usage strategies, though the margin is smaller for stronger models whose baseline performance is already high. These results suggest that allowing agents to decide when to consult distilled reasoning experiences is beneficial for knowledge-intensive reasoning.

Under SkillRL, we observe similar trends despite the different form of experience. Since SkillRL uses a pre-constructed skill repository, all variants share the same underlying experience memory and differ only in how skills are utilized at runtime. ExpWeaver consistently improves over vanilla usage on TriviaQA and HotpotQA, and remains competitive on 2Wiki. This provides controlled evidence that the gains arise from better runtime experience utilization rather than changes in the experience repository itself.
Together, these QA results complement the main results on ALFWorld and WebShop, showing that ExpWeaver generalizes beyond interactive environments to knowledge-intensive reasoning tasks.

\paragraph{Results on G-Memory.}
Figures~\ref{fig:g_memory_app} and~\ref{fig:g_memory_qa} present the prompting-based results under the G-Memory framework. In the G-Memory framework, we did not design a variant that retrieves experience at every step (Always-on), as doing so would violate its design philosophy and core architecture—namely, building a hierarchical graph memory store with complete task trajectories as the central nodes. In our experiments, we only adjusted the timing of experience presentation during execution, without changing G-Memory's own experience mechanism. This allows us to independently examine the effect of experience presentation timing while preserving the unique design advantages of the G-Memory framework.
Unlike the single-agent frameworks, G-Memory operates in a multi-agent setting and represents experience using both raw trajectories and distilled insights. Across ALFWorld, WebShop, and QA tasks, ExpWeaver generally improves over the vanilla initialization-only strategy, showing that interweaving experience utilization into decision-making remains effective in multi-agent self-evolving systems.

On ALFWorld, ExpWeaver consistently improves performance across all models. The gains are especially clear for Qwen3.5-397B-A17B and DeepSeek-V4-Pro, suggesting that even strong models benefit when experience is introduced at appropriate decision points rather than only at the beginning. On WebShop, ExpWeaver also improves over the vanilla strategy in most cases, indicating that adaptive experience usage helps agents resolve interactive decision uncertainty in web navigation.

For QA tasks, ExpWeaver achieves consistent or competitive improvements across TriviaQA, HotpotQA, and 2Wiki. The gains are generally smaller than those on ALFWorld, which is consistent with our usage-pattern analysis showing that QA tasks require less frequent experience invocation. Overall, the G-Memory results demonstrate that ExpWeaver generalizes beyond single-agent settings and remains effective when experience is shared and utilized in a multi-agent memory framework.

\paragraph{Results on AWM.}
Figures~\ref{fig:awm_app} and~\ref{fig:awm_qa} present the prompting-based results under the AWM framework. AWM represents experience as executable workflows and does not include an explicit retrieval mechanism during task execution. Therefore, the initialization-only and always-on variants are equivalent in implementation: the same workflow experience is provided as fixed guidance rather than dynamically retrieved at different steps.

Across ALFWorld, WebShop, and QA tasks, ExpWeaver generally improves over the vanilla AWM setting. The gains are especially clear on ALFWorld, where ExpWeaver consistently achieves higher success rates across different models, suggesting that even when experience is represented as workflows, allowing the agent to decide when to rely on experience remains beneficial. On WebShop and QA tasks, ExpWeaver also achieves competitive or improved performance in most settings.

These results further support our main conclusion: the benefit of ExpWeaver is not tied to retrieval-based memory alone. Even in a workflow-based framework without explicit retrieval, interweaving experience utilization into decision-making can improve how agents make use of existing experience.

\begin{figure*}
    \centering
    \subfigure[Checkpoint evaluation on NQ.]{\includegraphics[width=0.45\textwidth]{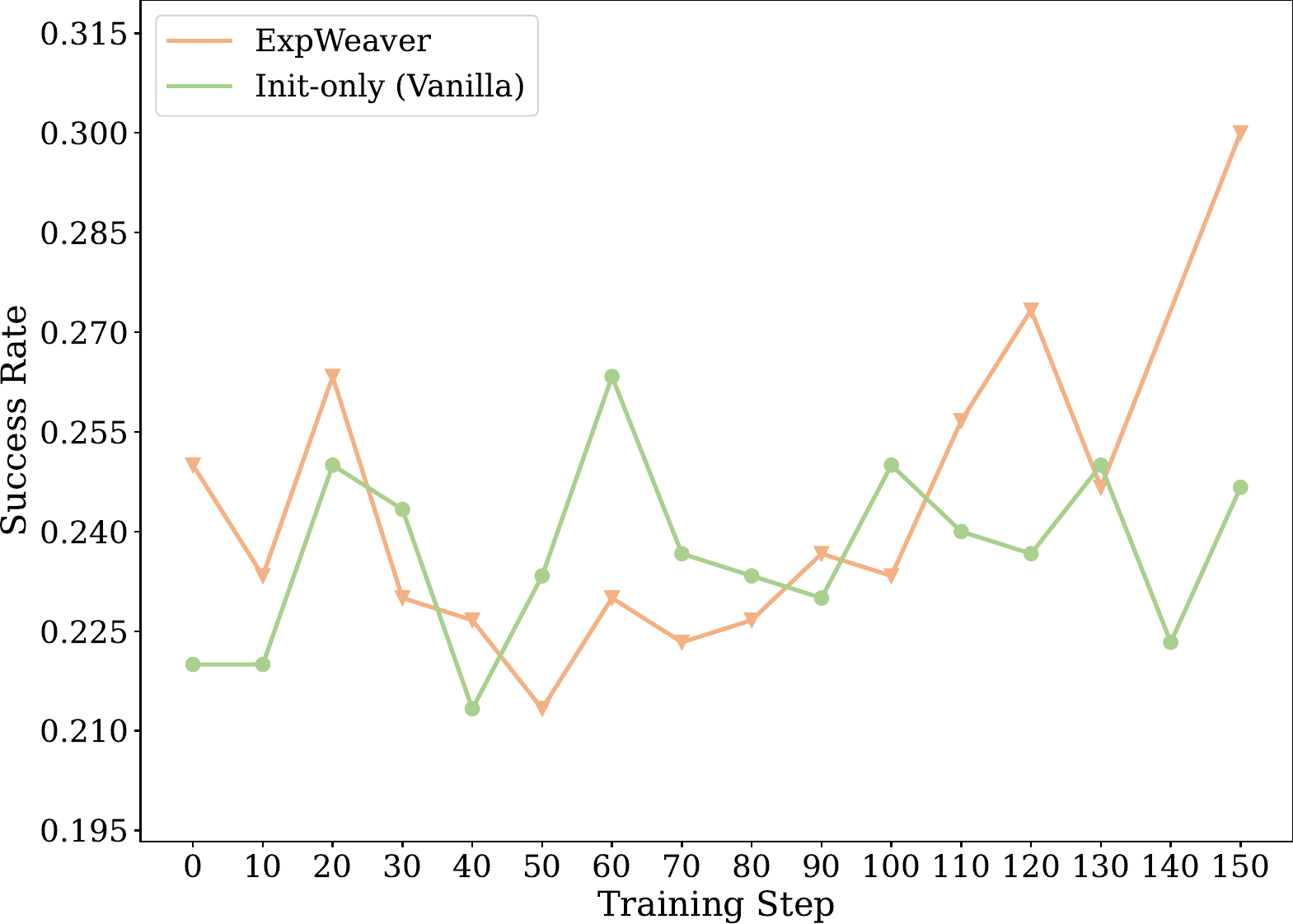}}
    \subfigure[Checkpoint evaluation on TriviaQA.]{\includegraphics[width=0.45\textwidth]{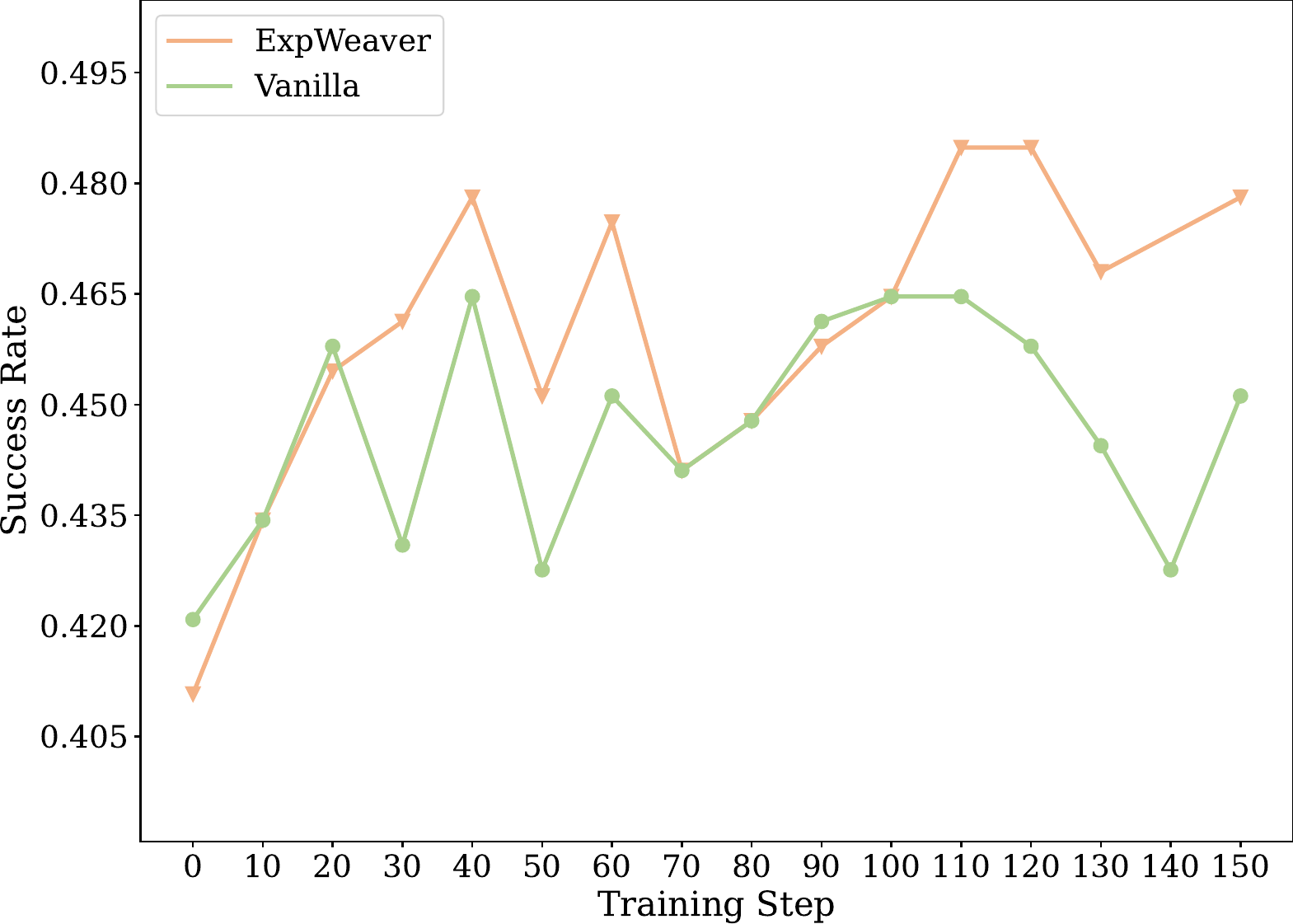}} \\
    \subfigure[Checkpoint evaluation on HotpotQA.]{\includegraphics[width=0.45\textwidth]{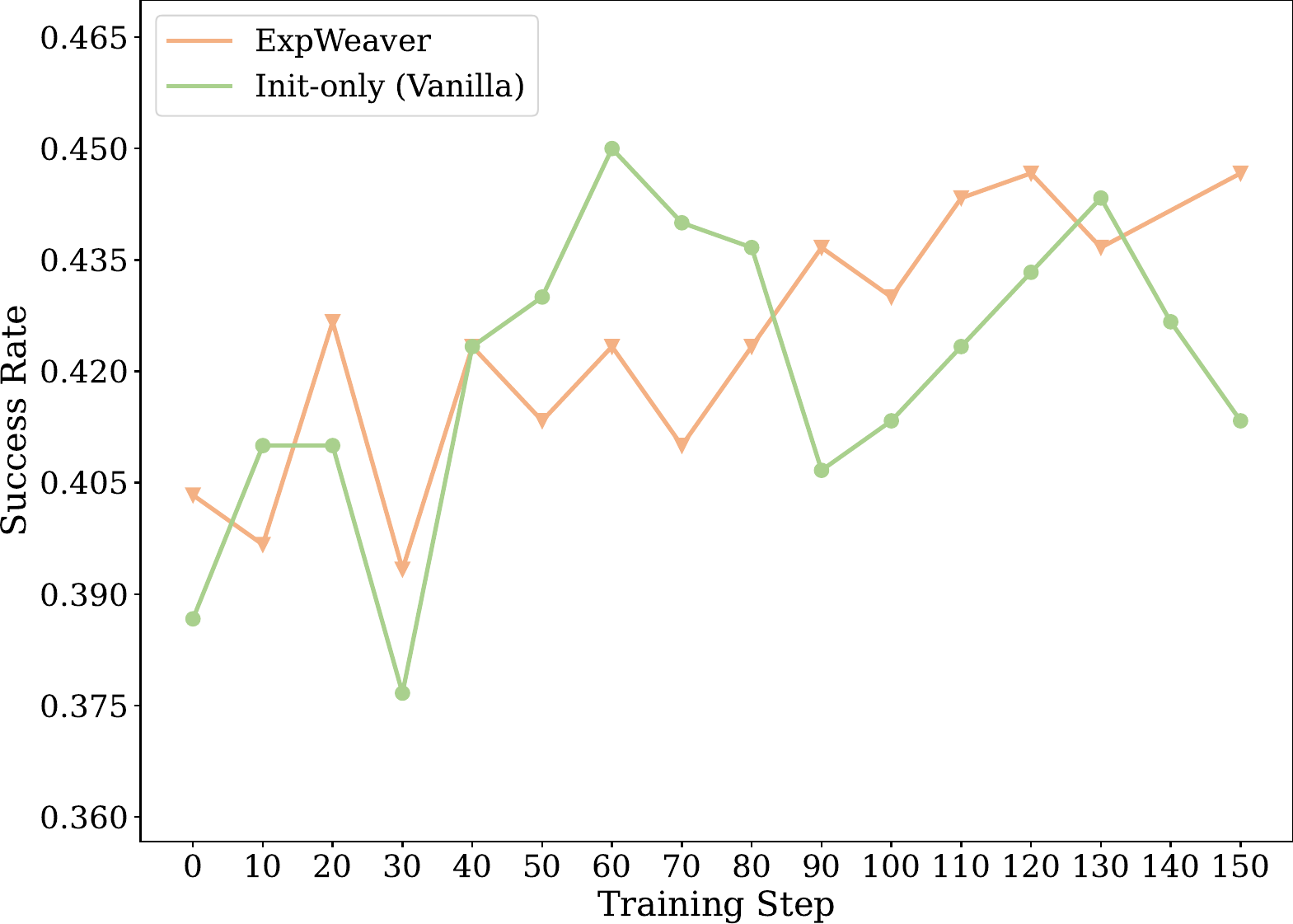}}
    \subfigure[Checkpoint evaluation on 2Wiki.]{\includegraphics[width=0.45\textwidth]{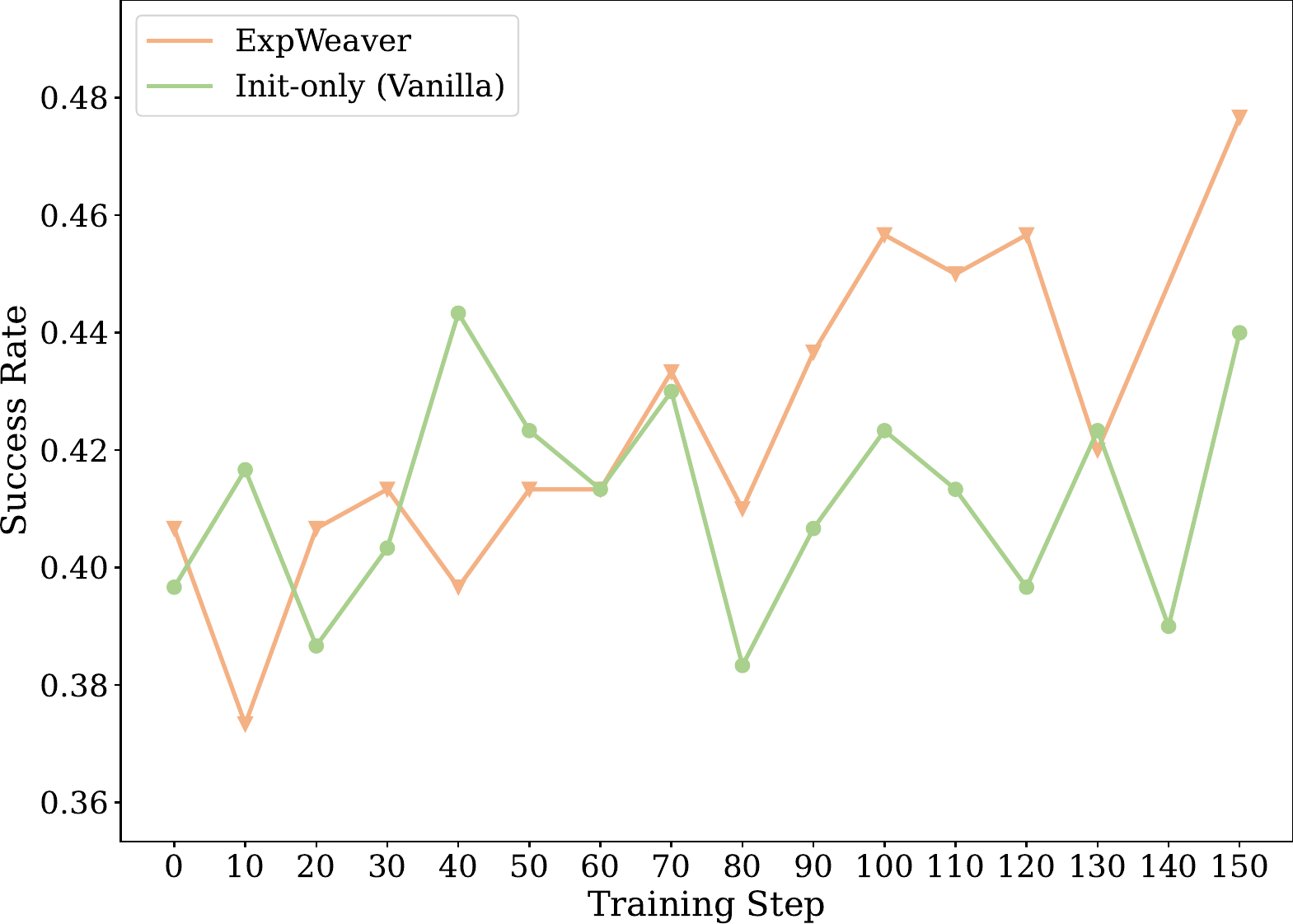}} \\
    \subfigure[Checkpoint evaluation on Bamboogle.]{\includegraphics[width=0.45\textwidth]{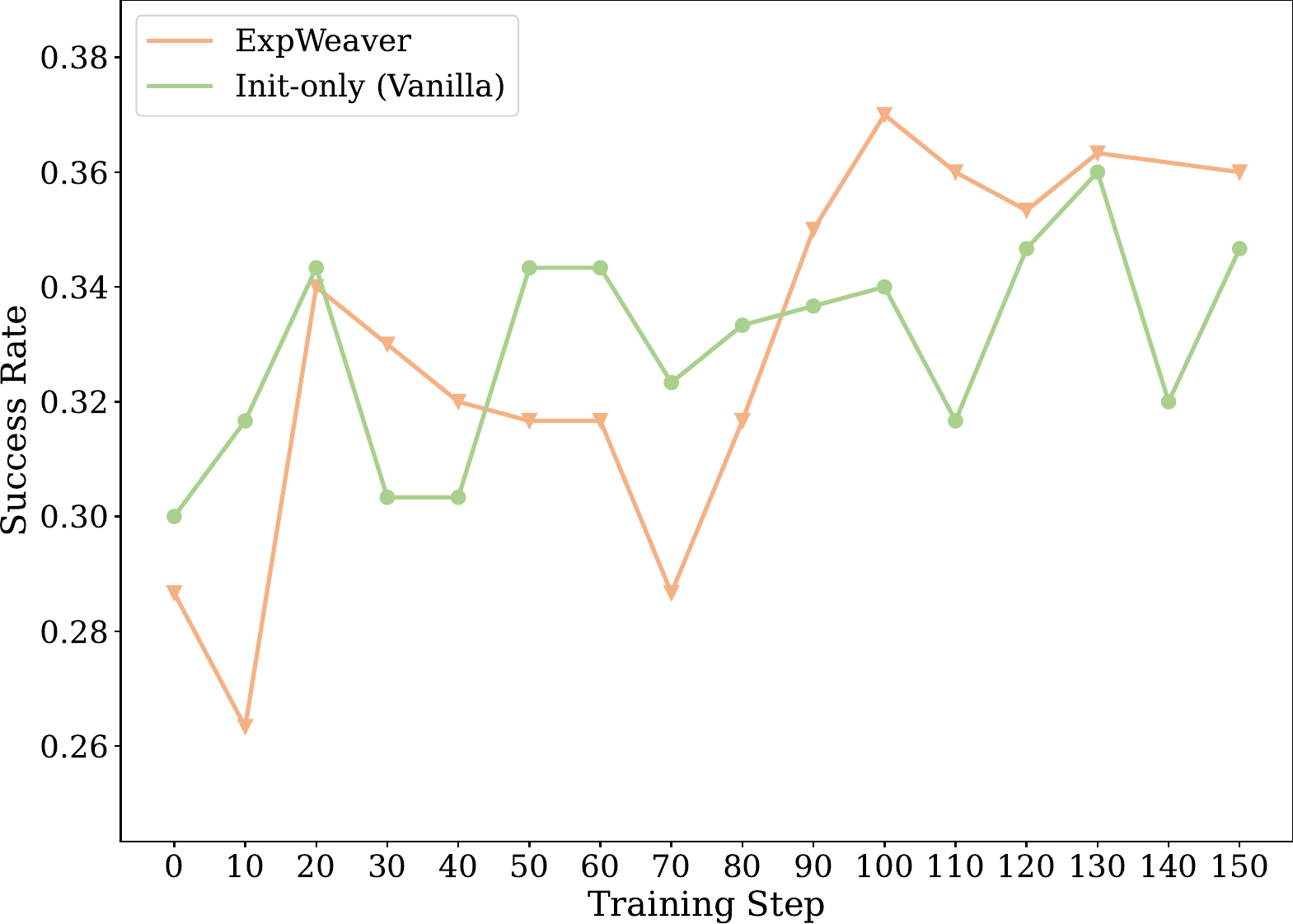}}
    \caption{Reinforcement learning results of Qwen3-14B on knowledge-intensive QA tasks.}
    \label{fig:rl_qa}
\end{figure*}

\subsection{Additional RL Results on QA Tasks}
\label{app:rl_qa}

We further evaluate RL-based ExpWeaver on knowledge-intensive QA tasks using Qwen3-14B under the ReasoningBank framework. Specifically, we train the model with GRPO on a merged training set consisting of NQ and HotpotQA \citep{jin2025search}. During training, we periodically evaluate intermediate checkpoints on five test sets: NQ, TriviaQA, HotpotQA, Bamboogle, and 2Wiki. Figure~\ref{fig:rl_qa} reports the success rates of ExpWeaver and the vanilla baseline across training steps, where each point corresponds to the performance of a checkpoint on the corresponding test set.

Overall, ExpWeaver achieves stronger improvements than the vanilla baseline (Init-only) across QA tasks. On NQ, ExpWeaver shows a clear upward trend and reaches a substantially higher success rate at later training steps, indicating that the model gradually learns to leverage experience more effectively during reasoning. Similar trends are observed on TriviaQA, Bamboogle, and 2Wiki, where ExpWeaver tends to achieve higher final performance.

These results suggest that the benefits of ExpWeaver are not limited to embodied interaction. Even in knowledge-intensive QA, where reasoning and evidence integration are central, allowing the agent to regulate when experience enters the reasoning process can improve RL optimization. This further supports our claim that interweaving experience utilization with decision-making is a learnable capability that generalizes beyond inference-time prompting and across task domains.

\section{Usage Pattern Analysis of ExpWeaver}

\begin{table*}[t]
\centering
\caption{
Additional usage pattern results on AWM and SkillRL. 
}
\label{tab:pattern_appendix}
\resizebox{\linewidth}{!}{
\begin{tabular}{lcccccc}
\toprule
\multirow{2}{*}{\textbf{Model}}
& \multicolumn{3}{c}{\textbf{AWM}}
& \multicolumn{3}{c}{\textbf{SkillRL}} \\

\cmidrule(lr){2-4} \cmidrule(lr){5-7}

& ALFWorld & WebShop & HotpotQA & ALFWorld & WebShop & HotpotQA  \\ 
\midrule
\texttt{Qwen3-32B} &2.62 &0.16 &0.53 &2.50 &0.14 &0.39 \\
\texttt{Qwen3.5-397B-A17B} &0.47 &0.00 &0.00 &0.28 &0.00 &0.02 \\
\texttt{DeepSeek-V4-Pro} &0.50 &0.58 &0.33 &0.60 &0.42 &0.11 \\
\texttt{GPT-5.2} &0.00 &0.00 &1.14 &0.00 &0.00 &0.00 \\
\texttt{Kimi-K2.5} &1.20 &0.00 &0.00 &0.88 &0.03 &0.08 \\
\bottomrule
\end{tabular}
}
\end{table*}

\subsection{Additional Horizontal Comparison  Results}
\label{app:usage_pattern}

Table~\ref{tab:pattern_appendix} reports the average number of experience retrievals per sample on AWM and SkillRL.

First, ALFWorld again induces substantially more experience retrievals than WebShop and HotpotQA. For example, under AWM, Qwen3-32B invokes experience 2.62 times per sample on ALFWorld, compared with 0.16 on WebShop and 0.53 on HotpotQA. Similarly, under SkillRL, Qwen3-32B retrieves experience 2.50 times per sample on ALFWorld, while retrieval frequencies remain below 0.40 on WebShop and HotpotQA. This reinforces the observation that embodied interaction, with long-horizon planning and partial observability, creates stronger demand for experience.

Second, stronger models tend to invoke experience less frequently. To be more specific, Qwen3.5-397B-A17B and GPT-5.2 exhibit near-zero retrieval on WebShop and HotpotQA under both AWM and SkillRL, suggesting that stronger backbones can often solve these tasks without additional experience. In contrast, smaller or less capable models rely on experience more often, especially in ALFWorld. This further supports the view that ExpWeaver adapts experience usage according to model capability rather than following a fixed retrieval schedule.

Third, the consistency across AWM and SkillRL indicates that the usage pattern is not specific to a particular form of experience. AWM represents experience as workflows, while SkillRL represents experience as reusable skills, yet both exhibit similar task- and model-dependent retrieval behavior. This suggests that interweaving experience utilization into decision-making induces a general regulatory mechanism over experience use, regardless of how experience is constructed or represented.

Together, these results reveal an important property of ExpWeaver: it does not simply reduce or increase experience usage uniformly. Instead, it produces a calibrated usage pattern conditioned on task difficulty, model capability, and decision context. This provides direct evidence that experience utilization should not be treated as an always-on operation; its value depends critically on when the agent chooses to invoke it.

\begin{figure*}[htbp]
    \centering
    \subfigure[On WebShop.]{\includegraphics[width=0.48\textwidth]{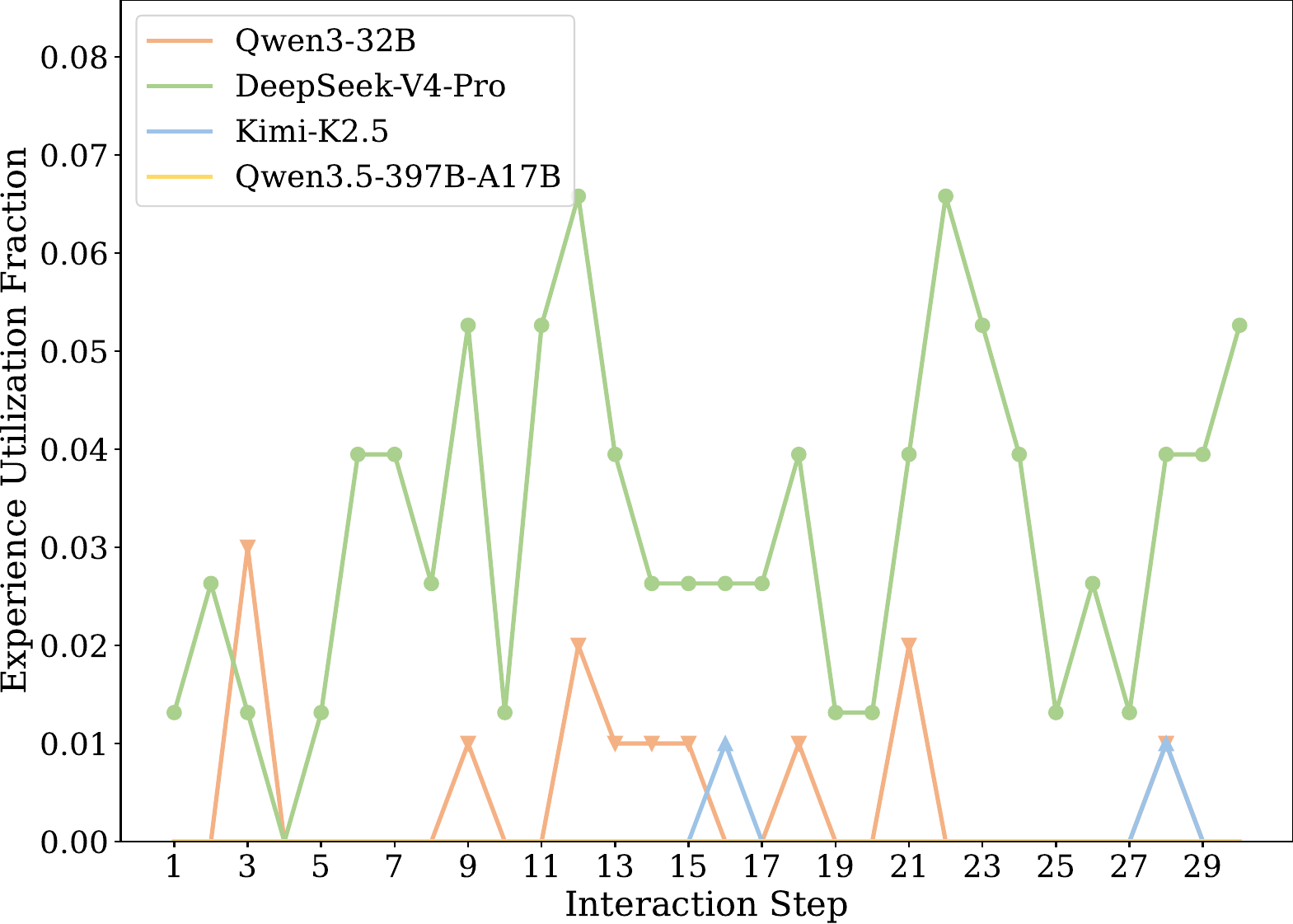}}
    \subfigure[On HotpotQA.]{\includegraphics[width=0.48\textwidth]{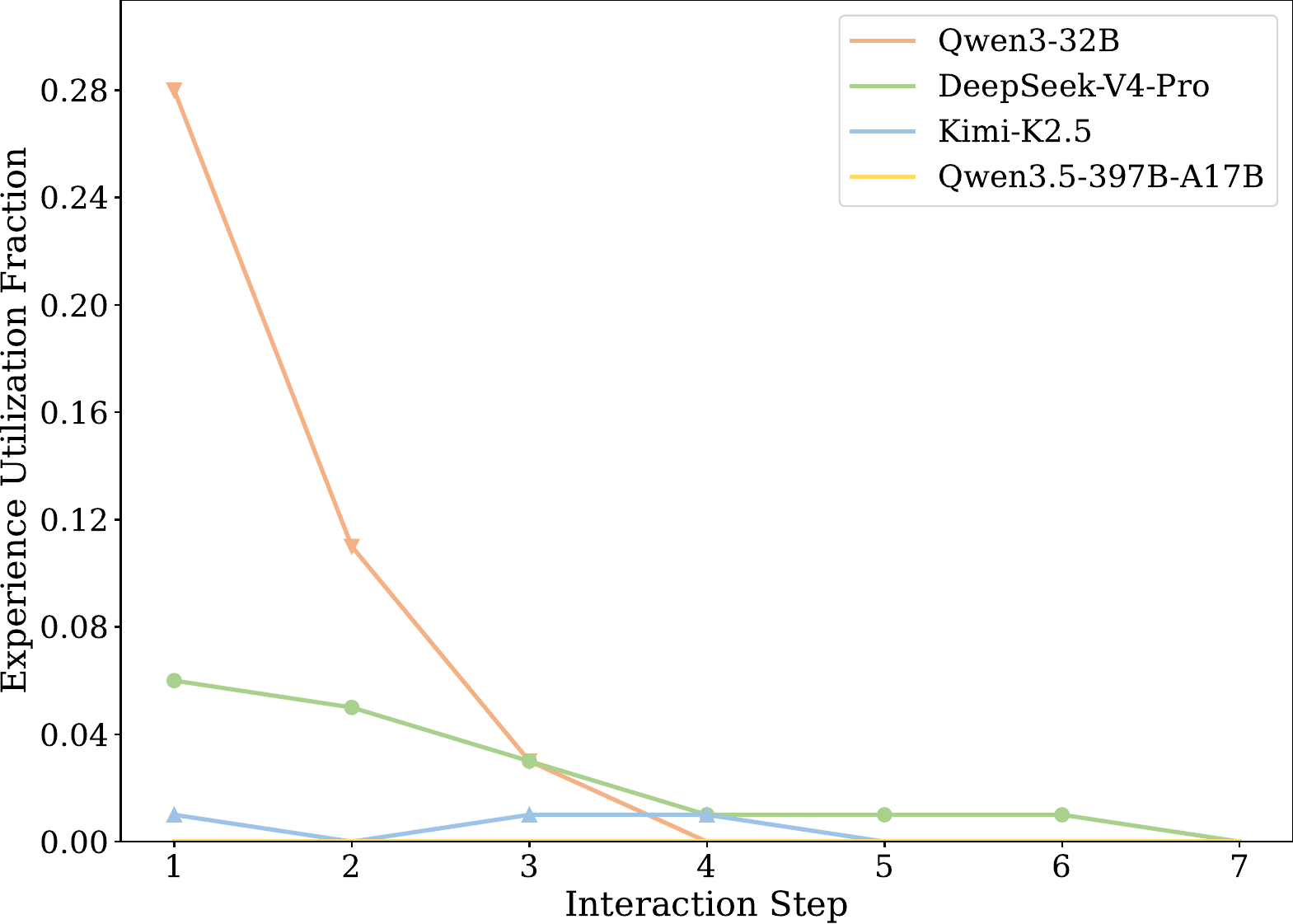}}
    \caption{Temporal pattern of experience utilization under ExpWeaver with ReasoningBank.}
    \label{fig:temporal_reasoningbank}
\end{figure*}

\begin{figure*}[htbp]
    \centering
    \subfigure[On ALFWorld.]{\includegraphics[width=0.45\textwidth]{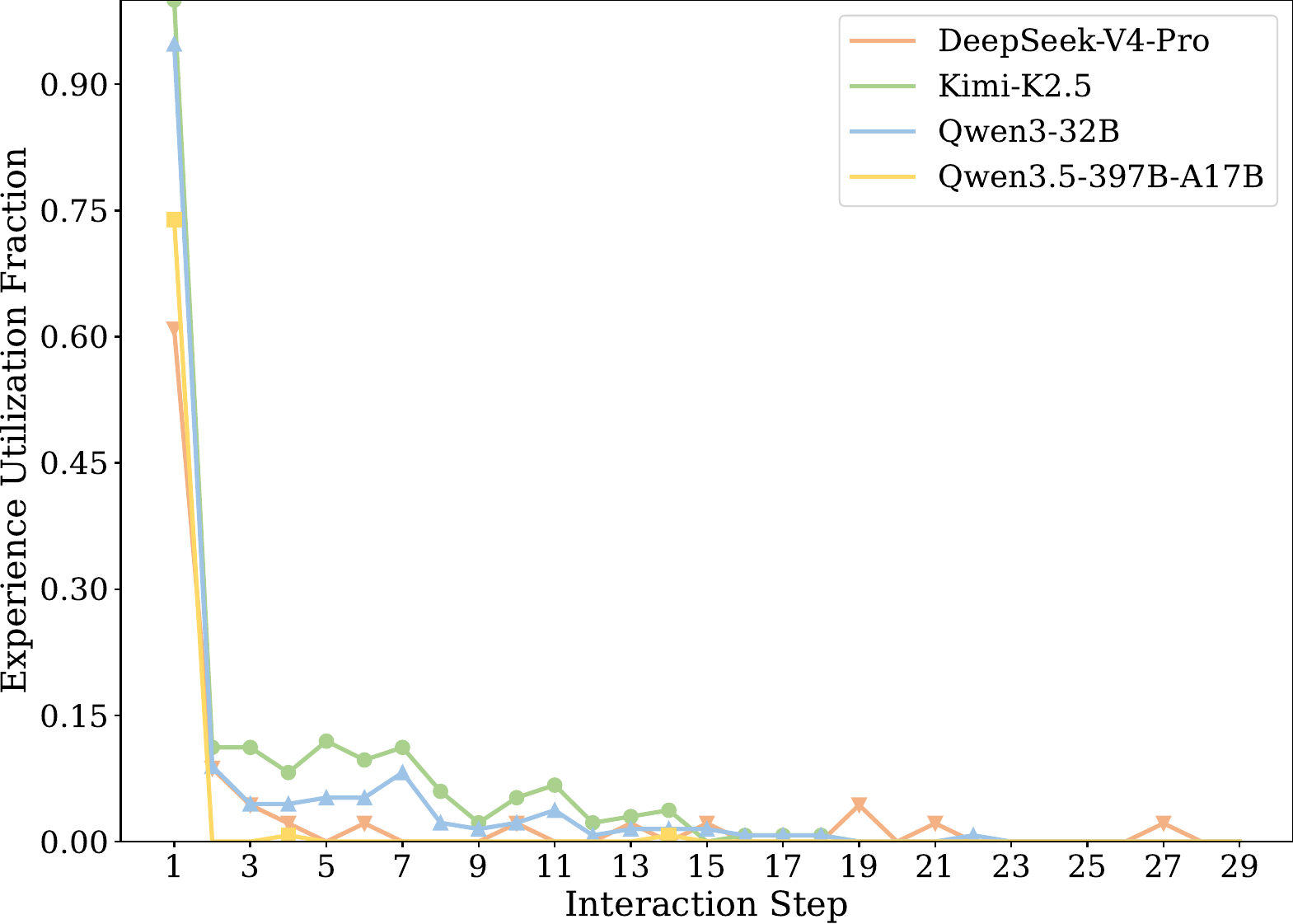}}
    \subfigure[On WebShop.]{\includegraphics[width=0.45\textwidth]{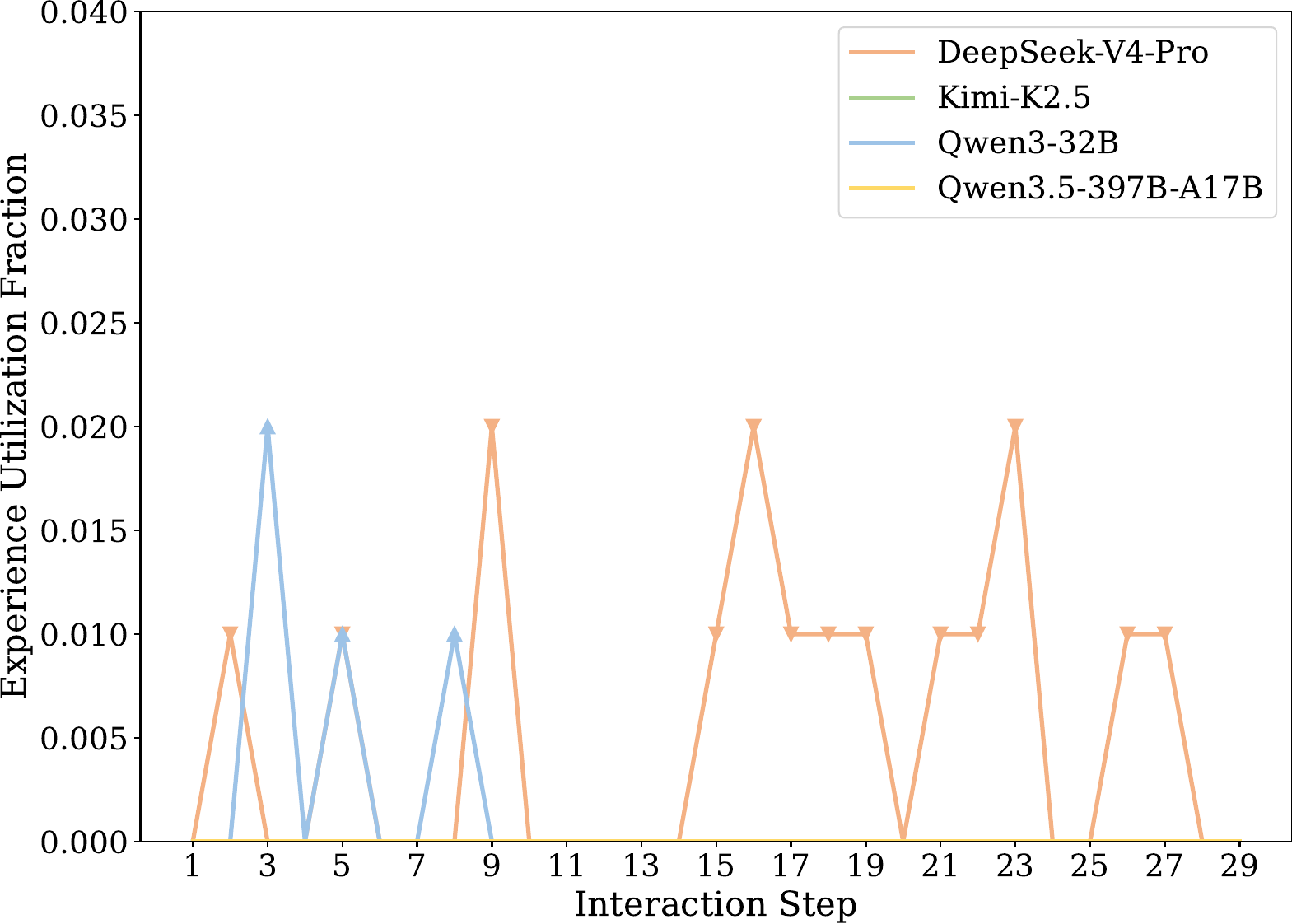}} \\
    \subfigure[On HotpotQA.]{\includegraphics[width=0.45\textwidth]{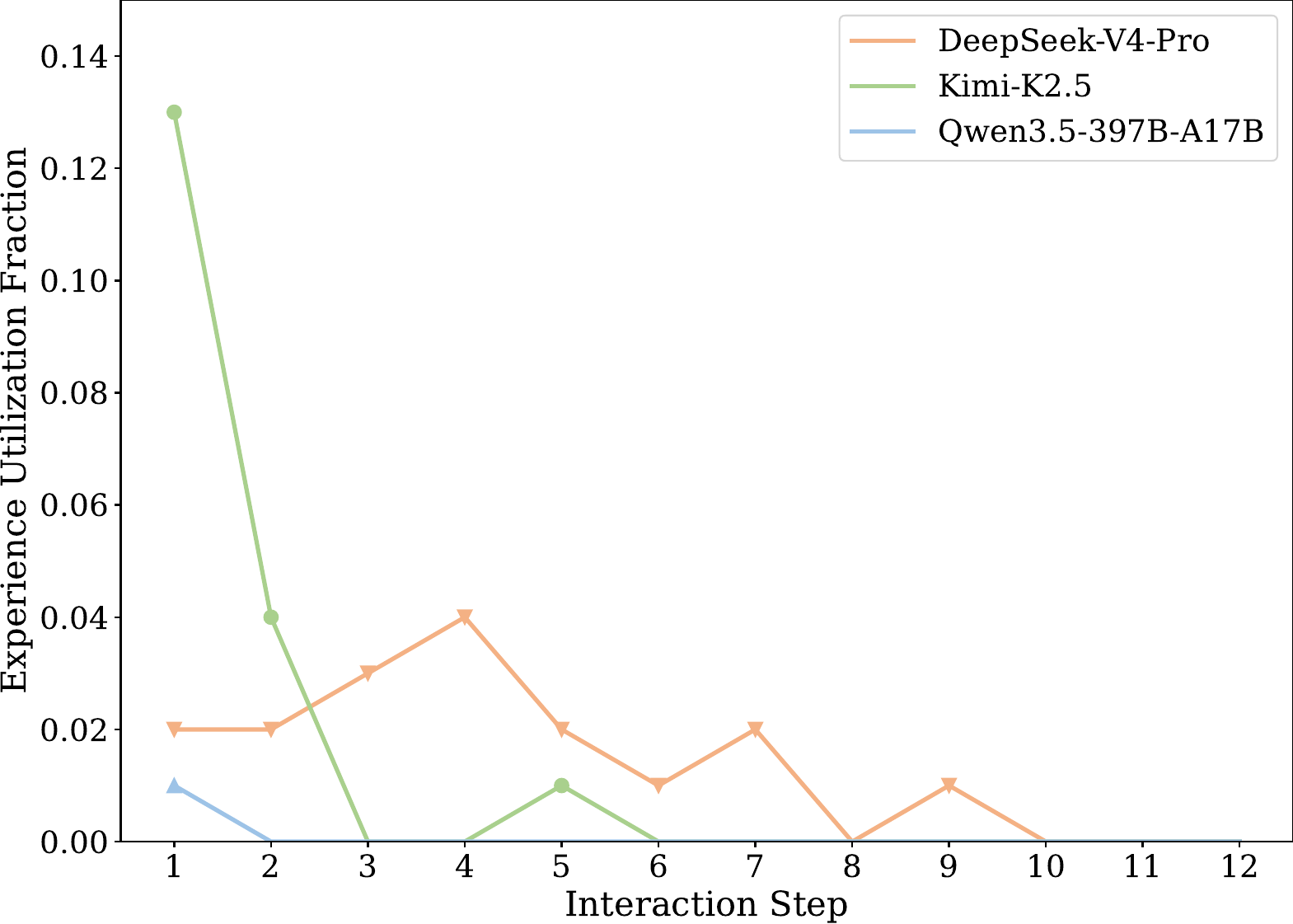}}
    \caption{Temporal pattern of experience utilization under ExpWeaver with G-Memory.}
    \label{fig:temporal_gmemory}
\end{figure*}

\begin{figure*}[htbp]
    \centering
    \subfigure[On ALFWorld.]{\includegraphics[width=0.45\textwidth]{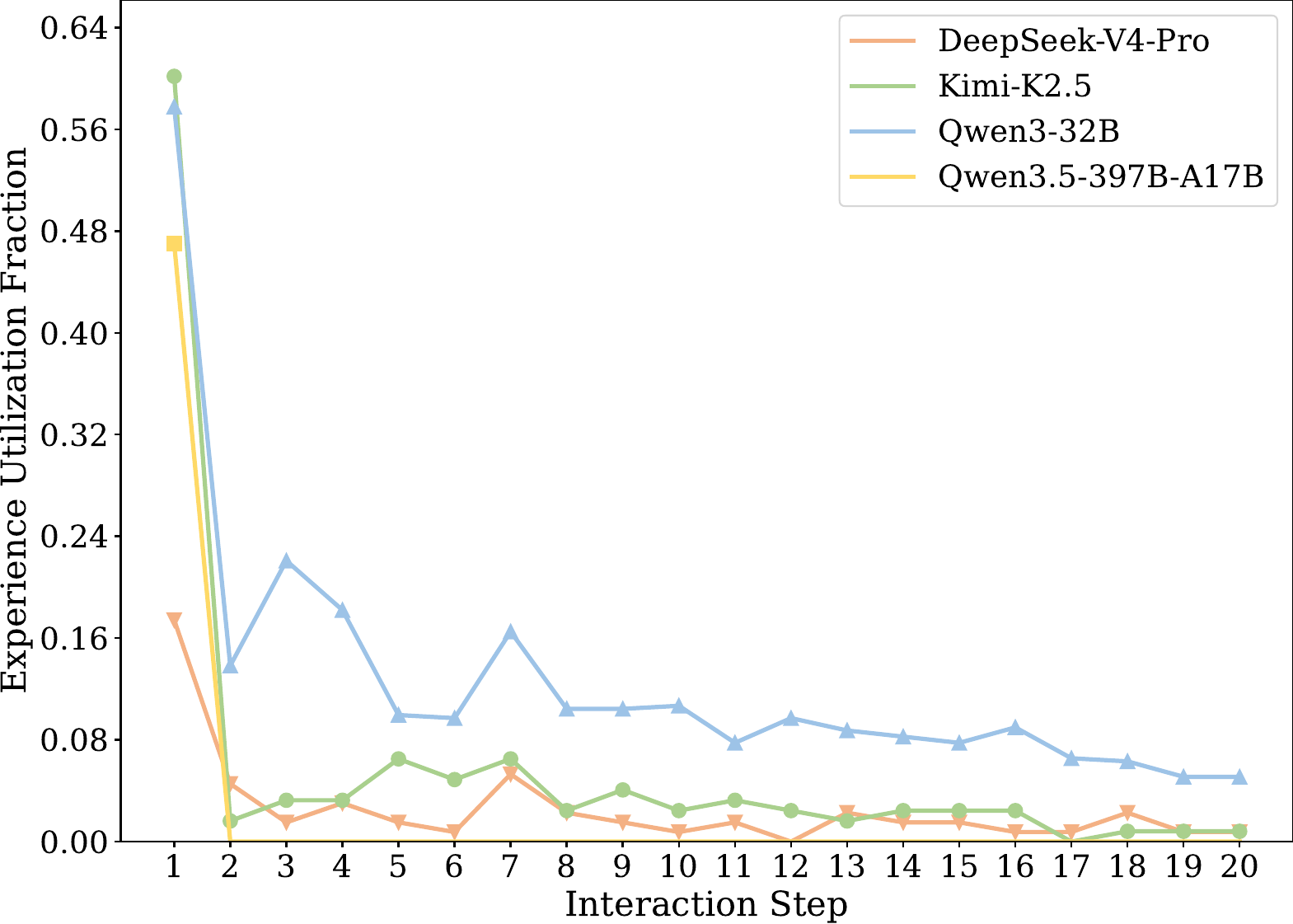}}
    \subfigure[On WebShop.]{\includegraphics[width=0.45\textwidth]{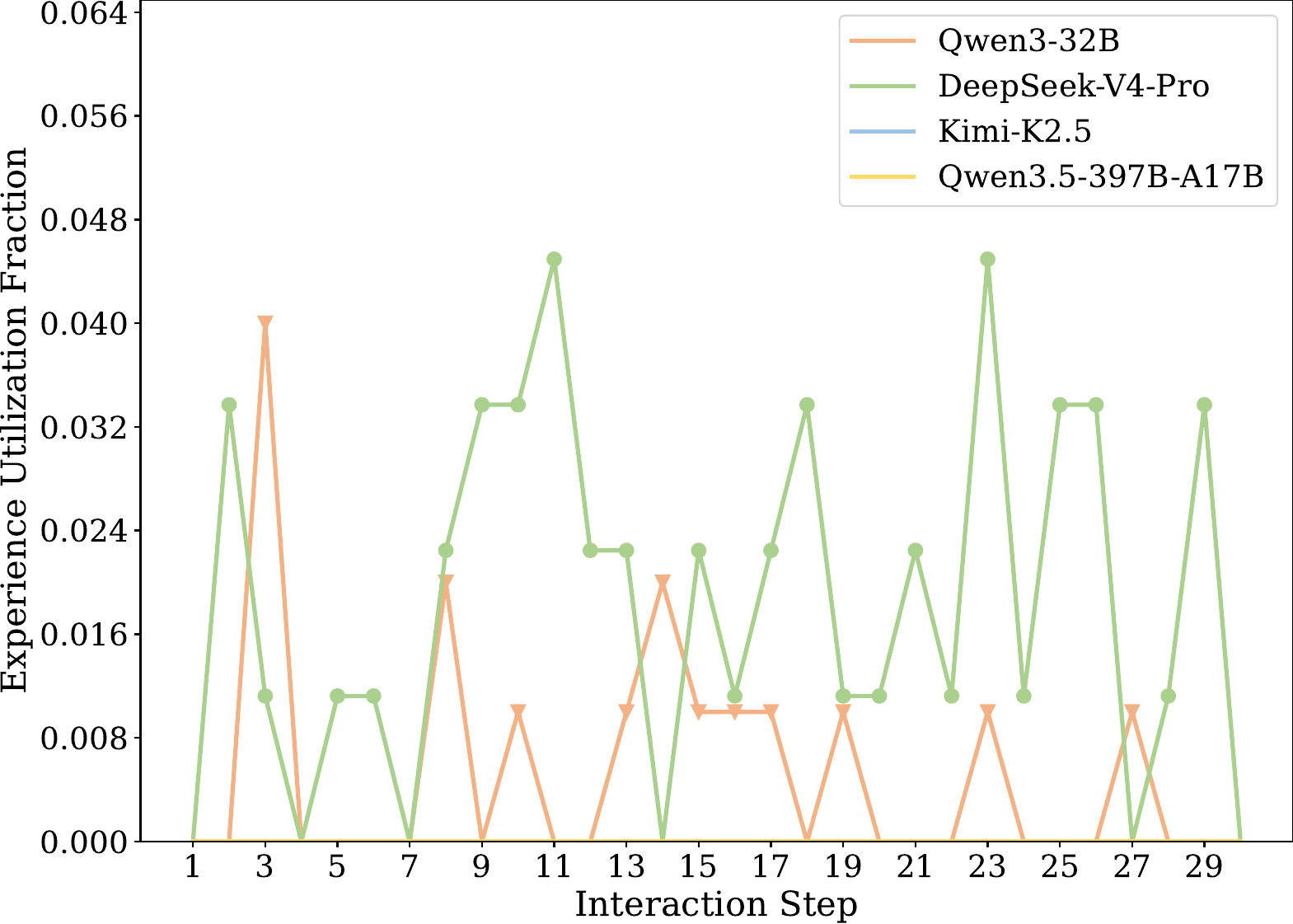}} \\
    \subfigure[On HotpotQA.]{\includegraphics[width=0.45\textwidth]{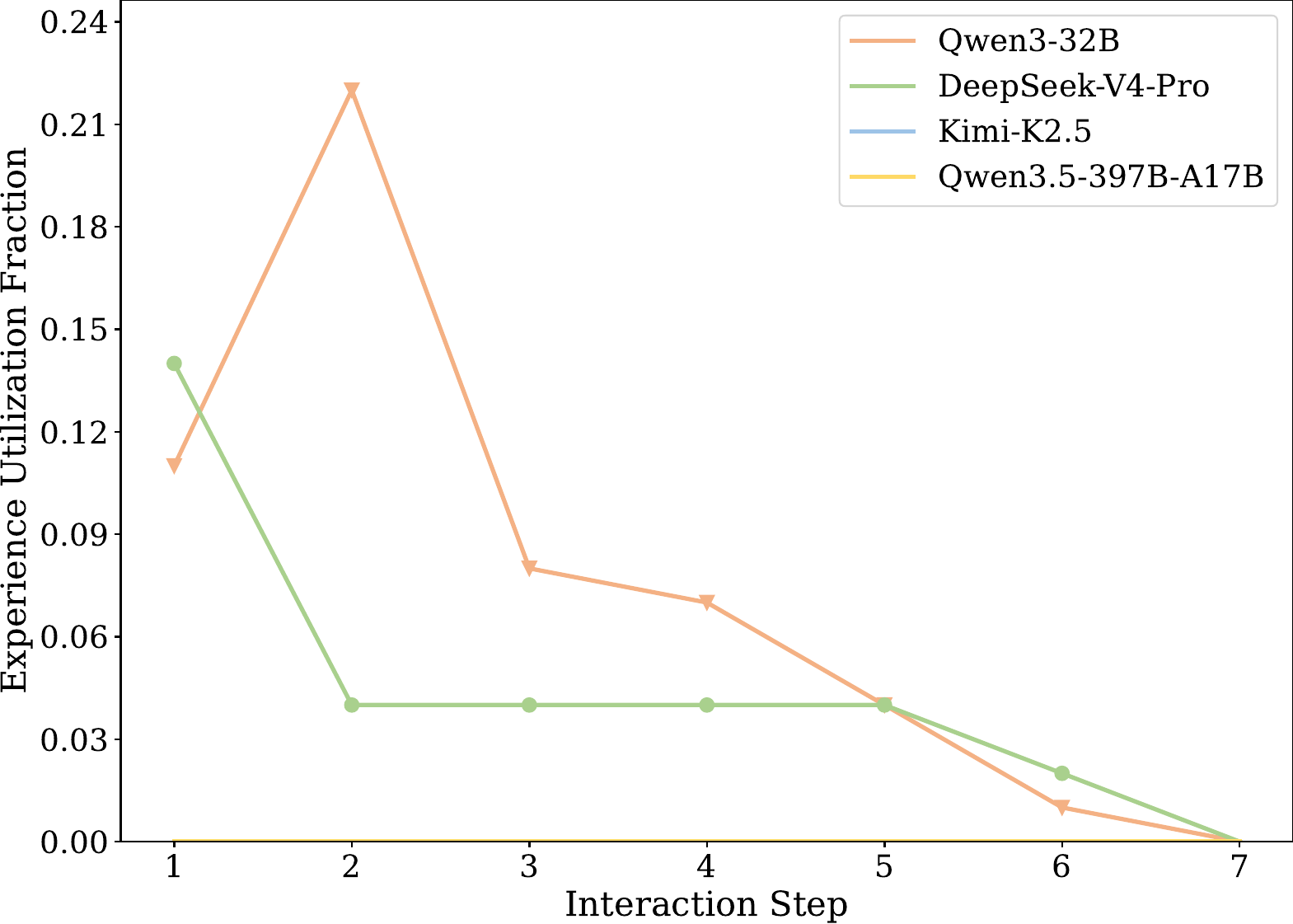}}
    \caption{Temporal pattern of experience utilization under ExpWeaver with AWM.}
    \label{fig:temporal_awm}
\end{figure*}

\begin{figure*}[htbp]
    \centering
    \subfigure[On ALFWorld.]{\includegraphics[width=0.45\textwidth]{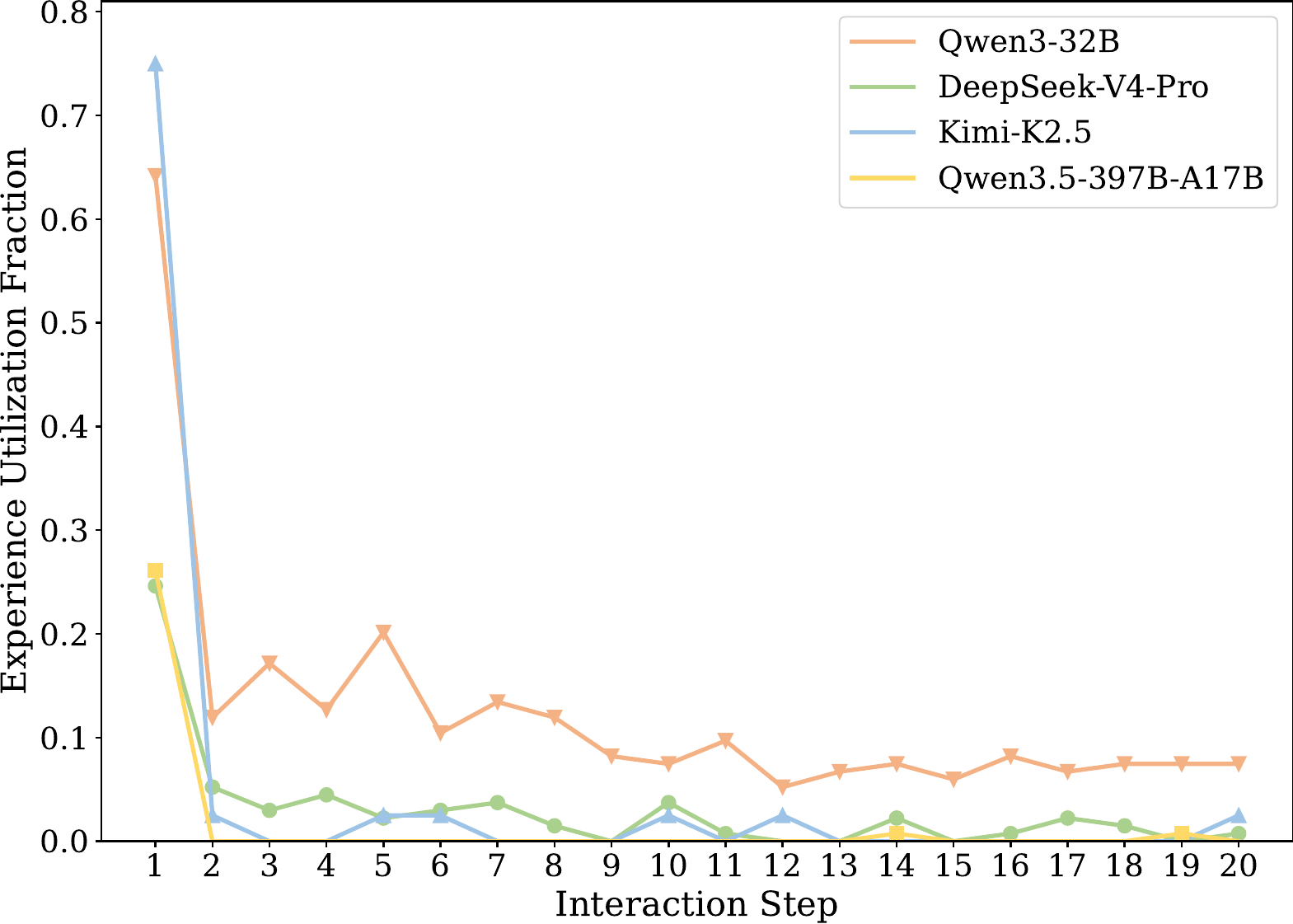}}
    \subfigure[On WebShop.]{\includegraphics[width=0.45\textwidth]{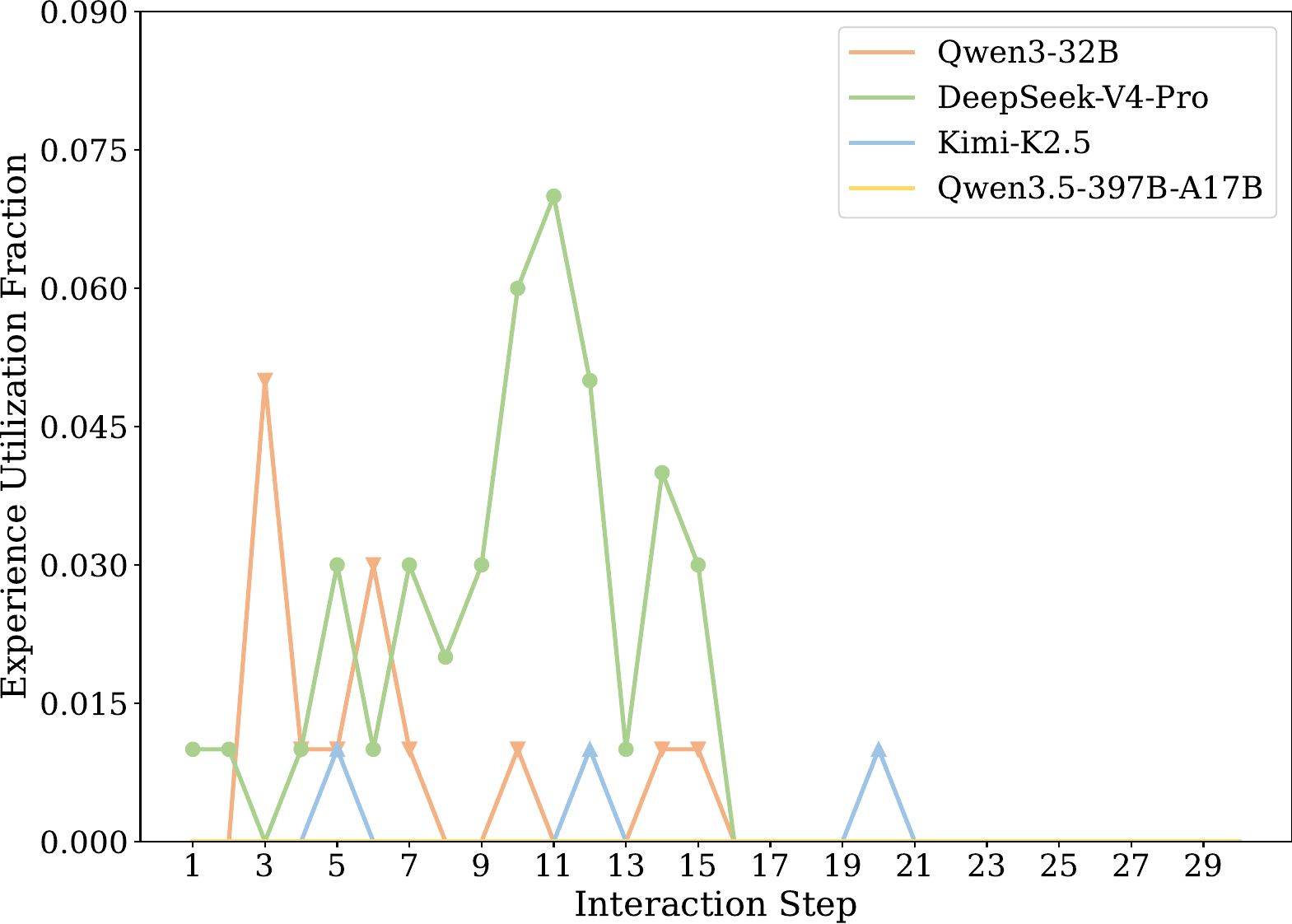}} \\
    \subfigure[On HotpotQA.]{\includegraphics[width=0.45\textwidth]{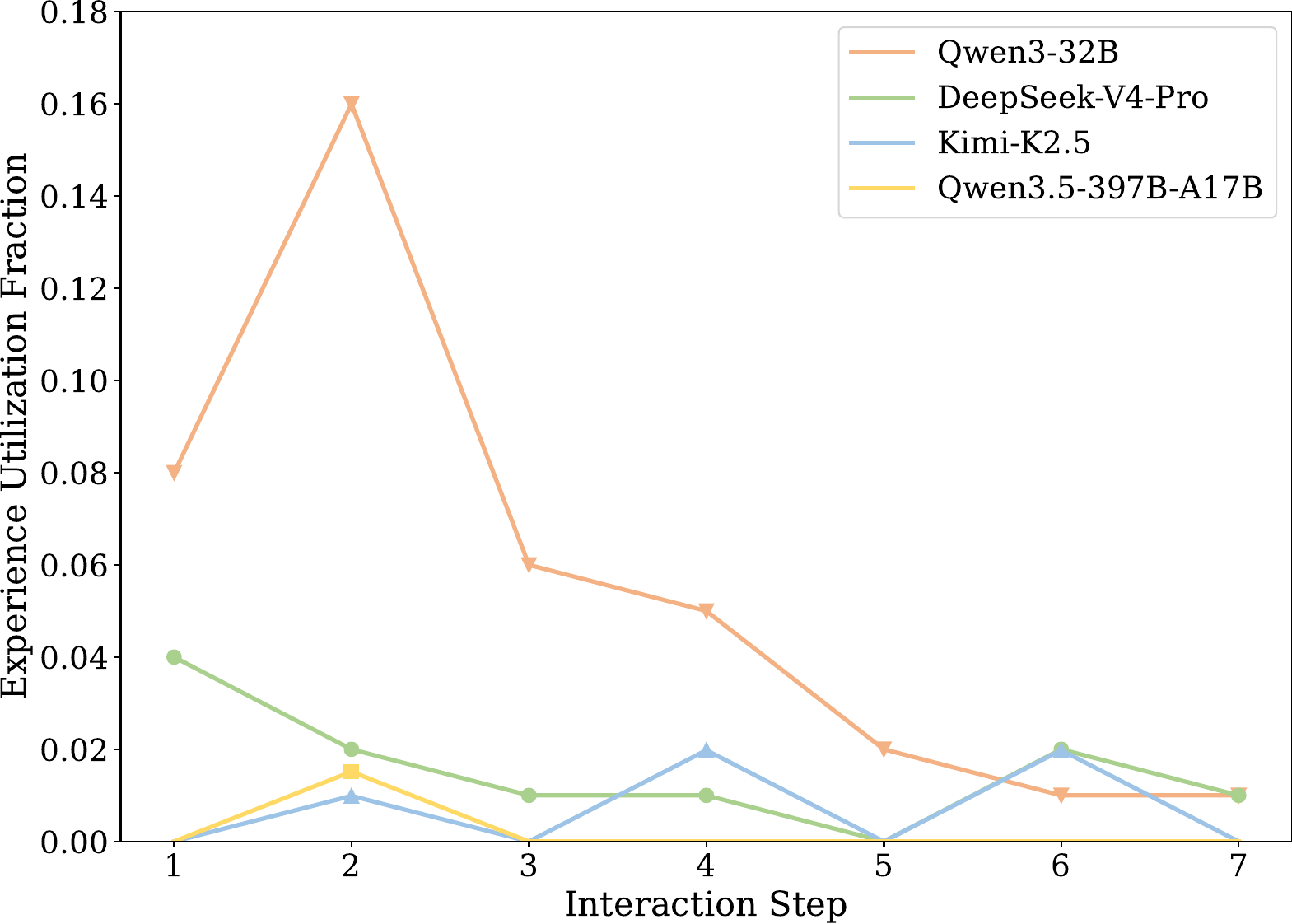}}
    \caption{Temporal pattern of experience utilization under ExpWeaver with SkillRL.}
    \label{fig:temporal_skillrl}
\end{figure*}

\begin{table*}[htbp]
\centering
\caption{The example of ExpWeaver on ALFWorld.}
\label{tab:alfworld_example}
\small
\setlength{\extrarowheight}{0pt}
\resizebox{\linewidth}{!}{
\begin{tabular}{p{0.9\linewidth}}
\toprule
\textbf{Task Goal}: heat some apple and put it in garbagecan. \\
\midrule
> think: I need to find an apple first. Apples are often found on diningtables, countertops, or in cabinets. Let me start by checking the diningtable.\\
> go to cabinet 1\\
You arrive at cabinet 1. On the cabinet 1, you see a glassbottle 1, and a soapbottle 1.\\
> think: No apple in cabinet 1. Let me check cabinet 2.\\
> go to cabinet 2\\
You arrive at cabinet 2. The cabinet 2 is closed.\\
> think: The cabinet 2 is closed. I need to open it to check for an apple.\\
> open cabinet 2\\
You open the cabinet 2. The cabinet 2 is open. In it, you see a plate 1.\\
> think: No apple found in cabinet 2 either. I have checked all visible locations without finding an apple. \textbf{I might need to recall where apples are typically found or consider other storage areas like the fridge. }\textbf{[RETRIEVE]}\\
<Retrieved Memory Items>\\
In past experiences, apples are often in the fridge or on the counter.\\
<End of Retrieved Memory>\\
> think: I have checked all visible locations without finding an apple. I might need to check the fridge next, as it is a common place to store apples. Let me go to the fridge.\\
> go to fridge 1\\
You arrive at fridge 1. The fridge 1 is closed.\\
> think: The fridge is closed. I need to open it to check for apples. Let me open the fridge.\\
> open fridge 1\\
You open the fridge 1. The fridge 1 is open. In it, you see a bowl 1, a bread 1, a cup 1, a lettuce 2, and a potato 1.\\
> think: No apple in the fridge either. I have exhausted all potential storage locations without finding an apple. \textbf{Perhaps I need to consider other possibilities or review past experiences to identify alternative search locations. }\textbf{[RETRIEVE]}\\
<Retrieved Memory Items>\\
The garbage can is in the kitchen. They should be checked.\\
<End of Retrieved Memory>\\
> go to garbagecan 1\\
You arrive at garbagecan 1. On the garbagecan 1, you see a apple 2, and a apple 1.
> think: I found apples in the garbagecan. I need to take one, heat it, and then return it to the garbagecan. Let me first take one of the apples.\\
> take apple 1 from garbagecan 1\\
You pick up the apple 1 from the garbagecan 1.\\
> think: Now that I have the apple, I need to heat it. The most reliable way to heat items is using the microwave. Let me go to the microwave.\\
> go to microwave 1\\
You arrive at microwave 1. The microwave 1 is closed.\\
> think: The microwave is closed. I need to open it before I can heat the apple. Let me open it.\\
> open microwave 1\\
You open the microwave 1. The microwave 1 is open.\\
> think: I'm at the microwave and have the apple. Let me heat the apple using the microwave.\\
> heat apple 1 with microwave 1\\
You heat the apple 1 using the microwave 1.\\
> think: The apple is now hot. The original plan might have been to place it on a plate after heating. I see a plate 1 inside cabinet 2. Let me go there and put the hot apple on the plate.\\
> go to cabinet 2\\
> You arrive at cabinet 2. The cabinet 2 is open. In it, you see a plate 1.\\
> put apple 1 on plate 1\\
> Task is SOLVED.\\
\bottomrule
\end{tabular}}
\end{table*}

\begin{table*}[htbp]
\centering
\caption{The example of ExpWeaver on WebShop.}
\label{tab:webshop_example}
\small
\setlength{\extrarowheight}{0pt}
\resizebox{\linewidth}{!}{
\begin{tabular}{p{0.9\linewidth}}
\toprule
\textbf{Task Goal}: I am looking for a long sleeve men t-shirt, color black, and price lower than 25.00 dollars. \\
\midrule
> Think[I need a black long sleeve men's t-shirt under \$25. I will search with the most specific attributes.]\\
> Search[long sleeve men t-shirt black]\\
> Observation:\\
{[}Back to Search{]}\\
{[}B09QQP3356{]}\\
HAUKLIE Men's Sports Waffle Ribbed Long Sleeve Henley T-Shirt\\
\$10.99\\

> Think[B09QQP3356 is a long sleeve men's shirt at \$10.99, well under \$25. Let me check its color options.]\\
> Click[B09QQP3356]\\
> Observation:\\
color [black][gray][light blue]\\
size [small][medium][large][x-large][xx-large]\\
HAUKLIE Men's Sports Waffle Ribbed Polo Shirts Summer Long Sleeve Cotton Muscle Quarter-Zip Henley T-Shirt\\
Price: \$10.99\\

> Think[The item has color and size options. The instruction requires black color. I will select black first, then pick a size. \textbf{I'm not sure whether I need to select size even when the instruction doesn't specify one. I'll search relevant experiences. [RETRIEVE]]}\\
{[}Retrieved Memory Items{]}\\
When an item page shows selectable options (size, color, flavor, etc.), always click each option to select it before clicking Buy Now. If the instruction does not specify a particular option value, select any available one (e.g., the first listed). Skipping option selection typically results in a wrong item or task failure.\\
{[}End of Retrieved Memory{]}\\

> Click[black]\\
> Observation: You have clicked black.\\

> Think[Black color selected. The instruction does not specify size, so I will choose the first available size.]\\
> Click[small]\\
> Observation: You have clicked small.\\

> Think[Both color and size are selected. Price \$10.99 is within the \$25.00 budget. Ready to buy.]\\
> Click[Buy Now]\\
> Observation: Your score (min 0.0, max 1.0): 1.0\\
\bottomrule
\end{tabular}}
\end{table*}

\begin{table*}[htbp]
\centering
\caption{The example of ExpWeaver on QA task.}
\label{tab:qa_example}
\small
\setlength{\extrarowheight}{0pt}
\resizebox{\linewidth}{!}{
\begin{tabular}{p{0.9\linewidth}} 
\toprule
\textbf{Task Goal}: What profession does Nicholas Ray and Elia Kazan have in common? \\
\midrule
> Thought 1: I need to search Nicholas Ray and Elia Kazan, find their professions, then find the profession they have in common.\\
> Action 1: Search[Nicholas Ray]\\
> Observation 1: Nicholas Ray (born Raymond Nicholas Kienzle Jr., August 7, 1911 – June 16, 1979) was an American film director, screenwriter, and actor best known for the 1955 film Rebel Without a Cause.\\
> Thought 2: \textbf{I have one side's professions, but I should retrieve a quick comparison checklist to avoid missing overlaps. [RETRIEVE]}\\
{[}Retrieved Memory Items{]}\\
Strategy: For profession-comparison questions, extract profession sets for both entities separately, then return only the intersection.\\
{[}End of Retrieved Memory{]}\\
> Thought 3: Following the checklist, I now need Elia Kazan's professions and then compute the intersection.\\
> Action 3: Search[Elia Kazan]\\
> Observation 3: Elia Kazan was an American film and theatre director, producer, screenwriter and actor.\\
> Thought 4: The intersection is director, screenwriter, and actor.\\
> Action 4: Finish[director, screenwriter, actor]\\
\bottomrule
\end{tabular}}
\end{table*}

\subsection{Additional Temporal Analysis}
\label{app:temporal_analysis}

We provide additional temporal analyses of experience utilization across different self-evolving frameworks and tasks. Figures~\ref{fig:temporal_reasoningbank}--\ref{fig:temporal_skillrl} report the fraction of trajectories that invoke experience at each interaction step under ReasoningBank, G-Memory, AWM, and SkillRL, respectively. Overall, these results show that the non-uniform temporal pattern observed in the main text is consistent across frameworks, tasks, and backbone models.

Across ALFWorld, experience utilization is highly concentrated at the beginning of trajectories. This trend appears consistently under G-Memory, AWM, and SkillRL, where most retrievals occur in the first few interaction steps and then rapidly decrease. This suggests that agents tend to consult experience when initial task understanding, high-level planning, and environment exploration are most needed. Notably, some models continue to invoke experience sporadically in later steps, especially in ALFWorld, which often corresponds to object-search or recovery phases under partial observability.

On WebShop, the temporal patterns are more dispersed and generally sparser. Compared with ALFWorld, agents invoke experience less frequently, and retrieval events are often scattered across later interaction steps. This reflects the different nature of web navigation, where the agent may not need experience at the beginning but may consult it when encountering uncertain product search, comparison, or navigation decisions. The scattered peaks indicate that experience is used as situational guidance rather than as a fixed initialization signal.

On HotpotQA, experience utilization is typically concentrated in early reasoning steps and quickly diminishes afterward. This is especially clear for models with higher retrieval frequency, where experience is often invoked during initial question decomposition or evidence-seeking. Once the agent establishes a reasoning direction, later retrieval becomes less frequent. This pattern supports the view that experience is mainly used to guide early reasoning or resolve uncertainty, rather than being repeatedly injected throughout the entire reasoning process.

Importantly, these trends hold across different experience representations, including distilled insights in ReasoningBank, raw trajectories and distilled insights in G-Memory, workflows in AWM, and reusable skills in SkillRL. Although the absolute retrieval frequency varies across frameworks and models, the temporal structure remains consistent: ExpWeaver invokes experience selectively, with strong early-stage usage and occasional later-stage reactivation when the agent faces uncertainty or difficulty. These results further support our claim that interweaving experience utilization with decision-making enables agents to regulate when experience enters the reasoning process, rather than following rigid initialization-only or always-on usage strategies.

\subsection{Qualitative Examples}
\label{app:qualitative_examples}

We provide qualitative examples to illustrate how ExpWeaver invokes experience during decision-making across different environments. These cases show that experience retrieval is not triggered uniformly, but is typically invoked when the agent encounters uncertainty, ambiguity, or repeated failure during task execution.

\paragraph{ALFWorld.}
Table~\ref{tab:alfworld_example} presents an example from ALFWorld, where the agent needs to heat an apple and place it in the garbage can. The agent first searches several plausible locations, such as cabinets and the fridge, but fails to find the target object. After repeated unsuccessful search attempts, ExpWeaver triggers experience retrieval to seek guidance about alternative object locations. The retrieved experience suggests that the garbage can should be checked, which directly helps the agent locate the apple and continue the task. This example shows that ExpWeaver invokes experience when the agent is stuck in a partially observable environment and needs corrective guidance.

\paragraph{WebShop.}
Table~\ref{tab:webshop_example} shows a WebShop example where the agent must purchase a black long-sleeve men's t-shirt under a price constraint. After finding a candidate item, the agent becomes uncertain about whether unspecified options such as size should also be selected before buying. ExpWeaver retrieves relevant experience indicating that all selectable options should be clicked before purchasing, even when some options are not specified in the instruction. Guided by this experience, the agent selects the required color, chooses an available size, and successfully completes the purchase. This case shows that experience is useful for resolving procedural uncertainty in web environments.

\paragraph{Knowledge-intensive QA.}
Table~\ref{tab:qa_example} provides an example from a QA task asking for the profession shared by Nicholas Ray and Elia Kazan. After retrieving information about one entity, the agent recognizes a risk of missing overlapping professions and invokes experience to obtain a comparison checklist. The retrieved experience reminds the agent to extract profession sets for both entities separately and return their intersection. Following this, the agent searches for the second entity and correctly computes the shared professions. This illustrates how ExpWeaver helps regulate reasoning in knowledge-intensive tasks by retrieving experience when the agent finds a reasoning pitfall.

Overall, these examples support our temporal and entropy-based analyses: ExpWeaver tends to invoke experience at moments where the agent faces uncertainty or requires additional guidance. The retrieved experience then serves as targeted support for the current decision, rather than as static context uniformly injected throughout the trajectory.

%%%%%%%%%%%%%%%%%%%%%%%%%%%%%%%%%%%%%%%%%%%%%%%%%%%%%%%%%%%%

% \newpage
% \input{checklist.tex}

\end{document}